%% file: main.tex
\documentclass[letterpaper, 10 pt, conference]{ieeeconf}  % Comment this line out if you need a4paper

\IEEEoverridecommandlockouts                              % This command is only needed if
\usepackage{graphicx}
\usepackage{amsmath}
\usepackage{amssymb}
\usepackage{breqn}
\usepackage{subcaption}
\usepackage{algorithm}
\usepackage{algorithmicx,algpseudocode}
\algrenewcommand\algorithmicindent{0.6em}
\usepackage{pgfplots}
\usepackage{booktabs, dcolumn}

\pgfplotsset{
every axis/.append style={
  axis line style={->}, % arrows on the axis
  xlabel near ticks,
  ylabel near ticks,
  legend style={font=\scriptsize},
  label style={font=\scriptsize},
  tick label style={font=\scriptsize},
  title style={font=\scriptsize}
  }
}
\pgfplotsset{compat=newest}
\pgfplotsset{plot coordinates/math parser=false}
\newlength\figureheight
\newlength\figurewidth

\usepackage{stork-hang}

\makeatletter
\let\NAT@parse\undefined
\makeatother
\usepackage[numbers,sort,compress]{natbib}
\usepackage{hyperref}
\usepackage[font=footnotesize]{caption}
\usepackage{xcolor,colortbl}
\title{\LARGE \bf%
Reinforcement Learning in Topology-based Representation for\\ Human Body Movement with Whole Arm Manipulation   
}

\author{Weihao Yuan$^{1}$, Kaiyu Hang$^{3}$, Haoran Song$^{1}$, Danica Kragic$^{2}$, Michael Y. Wang$^{1}$ and Johannes A. Stork$^{2}$ % <-this % stops a space
\thanks{$^1$ These authors are with the Hong Kong University of Science and Technology, Hong Kong, China. W. Yuan is with the Department of Electronic and Computer Engineering. H. Song is with the Department of Mechanical and Aerospace Engineering. M. Y. Wang is with the Department of Mechanical and Aerospace Engineering and the Department of Electronic and Computer Engineering.}%
\thanks{$^2$ J. A. Stork and D. Kragic are with the Centre for Autonomous Systems, EECS, KTH Royal Institute of Technology, Stockholm, Sweden}%
\thanks{$^3$ K. Hang is with the Department of Mechanical Engineering and Material Science, Yale University, New Haven, Connecticut, USA}%
}

\begin{document}

\maketitle
% % \thispagestyle{empty}
% % \pagestyle{empty}

%%%%%%%%%%%%%%%%%%%%%%%%%%%%%%%%%%%%%%%%%%%%%%%%%%%%%%%%%%%%%%%%%%%%%%%%%%%%%%%%

%%%%%%%%%%%%%%%%%%%%%%%%%%%%%%%%%%%%%%%%%%%%%%%%%%%%%%%%%%%%%%%%%%%%%%%%%%%%%%%%
\input{includes/abstract}
\input{includes/intro}
\input{includes/relatedwork}

\input{includes/methodology}
\input{includes/experiments}
\input{includes/conclusion}


%%%%%%%%%%%%%%%%%%%%%%%%%%%%%%%%%%%%%%%%%%%%%%%%%%%%%%%%%%%%%%%%%%%%%%%%%%%%%%%%

{\small
\bibliographystyle{IEEEtranN}
\bibliography{ref}
}
\end{document}

%% file: includes/abstract.tex
% !TEX root =  ../main.tex
\begin{abstract}
%
%1. state the problem
Moving a human body or a large and bulky object can require the strength of whole arm manipulation (WAM).
%2. say why it is interesting
This type of manipulation places the load on the robot's arms and relies on global properties of the interaction to succeed---rather than local contacts such as grasping or non-prehensile pushing.
%3. say what your solution achieves
In this paper, we learn to generate motions that enable WAM for holding and transporting of humans in certain rescue or patient care scenarios. We model the task as a reinforcement learning problem in order to provide a behavior that can directly respond to external perturbation and human motion. For this, we represent global properties of the robot-human interaction with topology-based coordinates that are computed from arm and torso positions. These coordinates also allow transferring the learned policy to other body shapes and sizes.
%4. say what follows from your solution.
For training and evaluation, we simulate a dynamic sea rescue scenario and show in quantitative experiments that the policy can solve unseen scenarios with differently-shaped humans, floating humans, or with perception noise. Our qualitative experiments show the subsequent transporting after holding is achieved and we demonstrate that the policy can be directly transferred to a real world setting.

%Hence we model the interaction with skeleton curves and leverage topology-based representations, Writhe matrix and Laplacian coordinates, to describe the interaction and formulate a deep reinforcement learning problem. We propose to use these topology representations, rather than raw image or position coordinates as the state input to the network and learn a policy which is stable and well generalized, and can be directly transferred to reality.

%
\end{abstract}

%% file: includes/intro.tex
% !TEX root =  ../main.tex

\section{INTRODUCTION}

\begin{figure}[]
\centering
\setlength\tabcolsep{1mm}
\begin{tabular}{cc}
\includegraphics[height=0.5\columnwidth, trim={1cm 0cm 1cm 0cm}, clip]{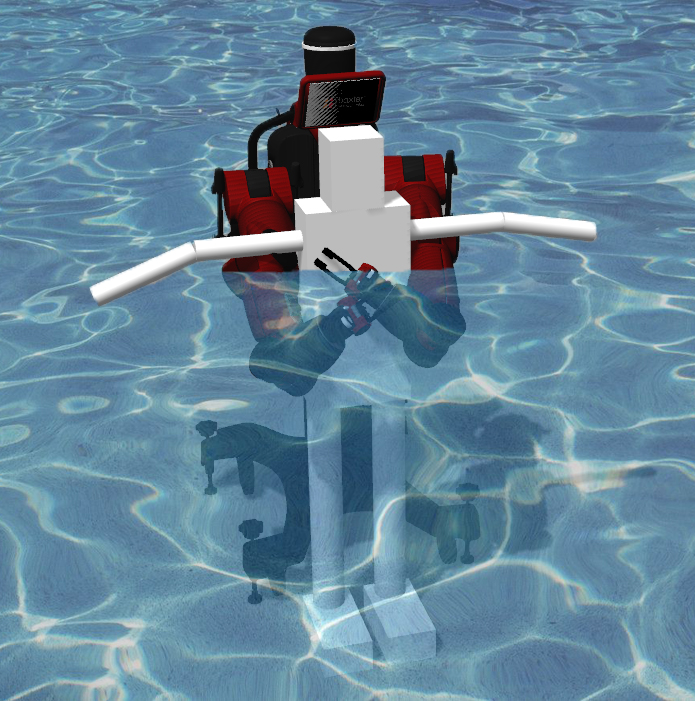}
&
\includegraphics[height=0.5\columnwidth]{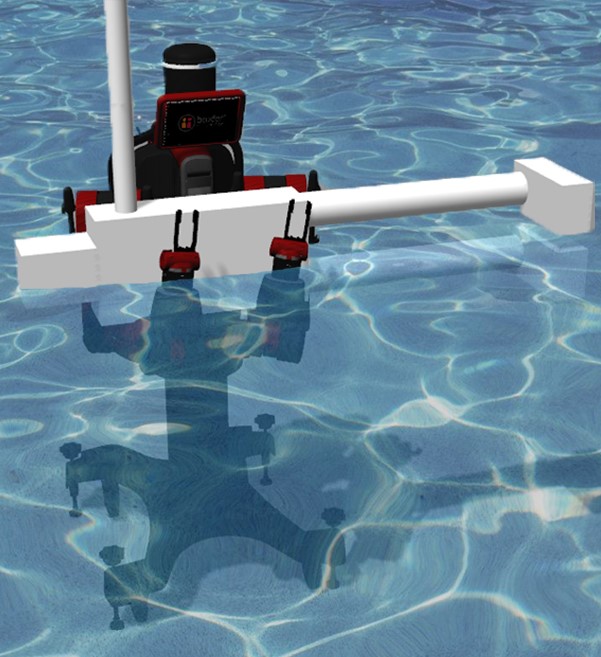}
\\
{\footnotesize (a) Upright} & {\footnotesize (b) Horizontal}
\end{tabular}
%
% \vspace*{-2mm}
%
\caption{%
%Swimming rescue is dangerous for human rescuer and a rescue robot could.
For a swimming rescue the robot has to firmly hold and then transport the drowning person which needs the strength of whole arm interaction. During the rescue, the person keeps move up and down due to waves and the robot has to continuously react to these changes.}
\label{fig:robot}
\end{figure}

Robotic manipulation is a complex problem that is often approached by grasping \cite{bicchi2000robotic, hang2017framework} or non-prehensile pushing \cite{cosgun11pushplanning, haustein2015kinodynamic, yuan2018rearrangement}. However, when heavy or bulky objects need to be manipulated \emph{whole arm manipulation} (WAM) is usually much more suitable \cite{nozawa2010full, florek2014humanoid, ohmura2007humanoid, onishi2007generation}. In WAM the robot's arms instead of its sensitive end-effectors are used to carry the load or provide support. This type of interaction is also often observed when somebody moves an injured person \cite{onishi2007generation} or rescues a drowning person at sea. Here, one or both arms are employed to embrace the person's body and then hold and transport the person as seen in Fig. \ref{fig:robot}.

In this paper, we learn to generate motions that enable WAM for holding and transporting of humans in certain rescue or patient care scenarios. This is a challenging WAM problem because humans have different sizes and shapes and can move and change their pose during interaction which is difficult to predict and model. Furthermore, a robot that is strong enough to move a person can easily cause injury.
WAM has previously been considered from the perspective of mechanical design \cite{salisbury1988preliminary, townsend1993mechanical}, robot control \cite{florek2014humanoid, song2000dynamic, bicchi1993force} and modeling of interaction \cite{watanabe2006towards}. In contrast to these works, we consider generating WAM-motions with a model-free learning-based approach and leave execution to a low-level controller.

Our scenarios require close interaction between the bodies of a humanoid robot and a person. This interaction is difficult to formalize for planning and control because of variation in geometry and uncertainty about physical response to contact forces.
Moreover, the success of this interaction depends on global properties which are difficult to determine geometrically, such as the form of entanglement between the two bodies.
Instead of referring to geometry, such as angles and positions of limbs, the magnitude of entanglement between limbs has therefore been considered for generating motions of two humanoid actors \cite{ho2010controlling, ho2011finite}. This topology-based representation called \emph{Writhe matrix} generalizes well to certain changes in body shape, size or their relative pose and we therefore employ it to capture the relationship of the two humanoid bodies in our scenarios.

Several works leverage Writhe matrix coordinates for generating motions: \citeauthor{ho2010controlling} interpolate between a set of key-poses and sequence more complex interactions with a state machine \cite{ho2010controlling, ho2011finite}, \citeauthor{stork2913integrated} use sampling-based planning to generate caging-grasps \cite{stork2913integrated, marzinotto2014cooperative}, and \citeauthor{ivan2013topology} present a control framework for motion planning \cite{ivan2013topology}. For our scenarios, these approaches are not flexible enough because they require defining the interaction using intermediate goals or do not continuously react to changes in the environment, such as waves during a swimming rescue. Instead, we employ model-free reinforcement learning to obtain a policy that can generate the desired motion.

In this context, we exploit the topology-based representation in two ways: Because of its invariance properties, we only have to train for one humanoid body shape and can apply the policy to humans of different shapes and sizes. Further, since the representation is based only on a simplified skeleton of the body, we can train in a virtual environment and apply the policy in reality without adaption as long as such a skeleton can be provided in the real scene.

\textbf{Our contributions} in this work are:
\begin{itemize}
\item formulating motion generation for WAM as a reinforcement learning problem and thus enabling reactive behavior,
\item exploiting Writhe and Laplacian coordinates in reinforcement learning of WAM interaction with humans,
\item modeling of two different dual-arm scenarios: interaction with \emph{upright} and \emph{horizontal} humanoid.
\end{itemize}

Our evaluation shows that we can reliably learn a policy that can generate the desired motion for different scenarios with a high success rate of $99\%$. In evaluation with humanoid bodies of different shapes and sizes, bodies in continuous motion, and artificial perception  noise, the policy still performs well. Additionally, we show a proof-of-concept for applying the policy in reality with a real robot and person.

%In our evaluation and test experiments, the robot performs well and stably, reaching a tight holding defined empirically within a few movements with a high success rate $99\%$, which indicates the effectiveness of our method. And the policy learned on a standard-figure humanoid can be directly applied to slim and fat humanoid, or even a humanoid floating in the air. This proves the good generalization of our approach. Or these various scenarios are the same to the policy in topological representation.

%% file: includes/relatedwork.tex
% !TEX root =  ../main.tex

\section{RELATED WORK}

In this section, we first review the works where robots use their arms to hold heavy or bulky objects and then survey learning methods that are similar to our approach.

The classical approach to manipulating bulky objects is based on physical modeling. For instance, \citeauthor{kaneko2000hugging} analyze forces and moments between the robot's legs and the object in order to maintain static balance \cite{kaneko2000hugging}. Similarly, \citeauthor{florek2014humanoid} propose an impedance controller to use contacts at both arms and the robot's chest to grasp a large object \cite{florek2014humanoid}. Different to these works, we do not consider contact forces since these are difficult to model for WAM interaction with humans. Instead we are interested in the spacial relationship between robot and human.

\citeauthor{marzinotto2014cooperative} maximize Writhe between robot arms and a tunnel hole in the object for collaborative grasping and transport of a large object \cite{marzinotto2014cooperative}. The representation and task formulation is similar to other works where Writhe or Linking is considered for caging grasps \cite{stork2013topology, pokorny2013grasping, stork2913integrated}, motion planning through holes \cite{zarubin2012hierarchical, ivan2013topology}, or animation of humanoid characters \cite{ho2010controlling, ho2010spatial}.
Similar to these works, we employ topology-based coordinates and aim to maximize the linking value between the robot and the person to reach a starting pose for transport. However, instead of sampling-based planning or optimal control which are time-consuming and not suitable for dynamic scenarios, we use reinforcement learning to find a policy which maximizes the linking value.

%This task can work in dynamic environment but it is not generalized to various scenarios with different objects. Similarly, \citeauthor{stork2013topology} \cite{stork2013topology, pokorny2013grasping} \cite{marzinotto2014cooperative} address the grasping, clasping and hooking manipulation problems with topology representation but their environments are all static.

%Ivan and Zarubin et al. \cite{zarubin2012hierarchical, ivan2013topology} combine the topology representations and other classical representations such as joint configuration and end-effector position to form a complex synthesis system to control the robot arm moving avoiding the wall.
%Ho et al. \cite{ho2010controlling, ho2010spatial} use topology coordinates to control the robots to do wrestling games, where the winding between the wrestlers is evaluated with writhe and intermesh. But because they use a finite state machine to control the motion of the robots, the states in this task are fixed and hard to apply.

Since deep reinforcement learning has shown success in complex artificial domains \cite{mnih2015human, silver2017mastering}, controlling robots with reinforcement learning has become increasingly interesting \cite{lillicrap2015continuous}. For instance, it has been used to learn grasping \cite{yahya2017collective, mahler2017dex} or manipulation in dynamic environments \cite{yuan2018rearrangement, OpenAI2018}. While these works exploit the advantages of deep models for visual input, this makes it difficult for them to generalize to different conditions. In contrast to that, we use topology-based coordinates as input to our policy. These coordinates are an abstraction for the actual shape and appearance of the robot and human and therefore intrinsically allow for generalization to different shapes and sizes.

%% file: includes/methodology.tex
\section{TOPOLOGICAL REPRESENTATION}

In this section, we describe how we represent the robot-humanoid relationship for our WAM scenario. This representation serves as input to the reinforcement learning policy described in Sec. \ref{sec:policy_learning}. We employ the concepts of Writhe matrix and Laplacian coordinates which we explain in Sec. \ref{sec:writhe} and Sec. \ref{sec:laplace} before we define our representation in Sec. \ref{sec:representation}.

\subsection{Writhe Matrix}
\label{sec:writhe}

The Writhe matrix $W$ with the entries $W_{i,j}$ is a representation of how much two curves, $\gamma_1$ and $\gamma_2$, wind around each other in three-dimensional space \cite{ho2010controlling}. While the Gaussian linking integral $\Gamma(\gamma_1, \gamma_2)$ represents this property as a single scalar \cite{pohl1968self},
%essentially counting the number of full
%
\begin{equation}
\Gamma(\gamma_1, \gamma_2)
 =
\frac{1}{4\pi}\int_{\gamma_1}\int_{\gamma_2}
\frac{d\gamma_1\times d\gamma_2 \cdot (\gamma_1-\gamma_2)}
{\|\gamma_1-\gamma_2\|^3}
\label{eq:gli}
,
\end{equation}
%
%where $\times$ and $\cdot$ represent cross product and dot product, respectively,
the Writhe matrix records this information separately for different segments of the two curves. For this, both curves are approximated with two sequences of line segments, indexed by $i = 1, 2, \dots n_1$ and $j = 1, 2, \dots n_2$, respectively. The entries of the Writhe matrix $W_{i, j}$ are defined for pairs of segments,
\begin{equation}
W_{i, j}
=
\Gamma(s_1^i, s_2^j)
,
\quad \forall i \forall j
,
\end{equation}
where $s_1^i$ and $s_2^j$ are line segments of the two curves. %For the special case of line segments, $\Gamma(s_1^i, s_2^j)$ has a closed form \cite{klenin2000computation}.

\begin{figure}[]
\centering
\includegraphics[width=0.6\columnwidth]{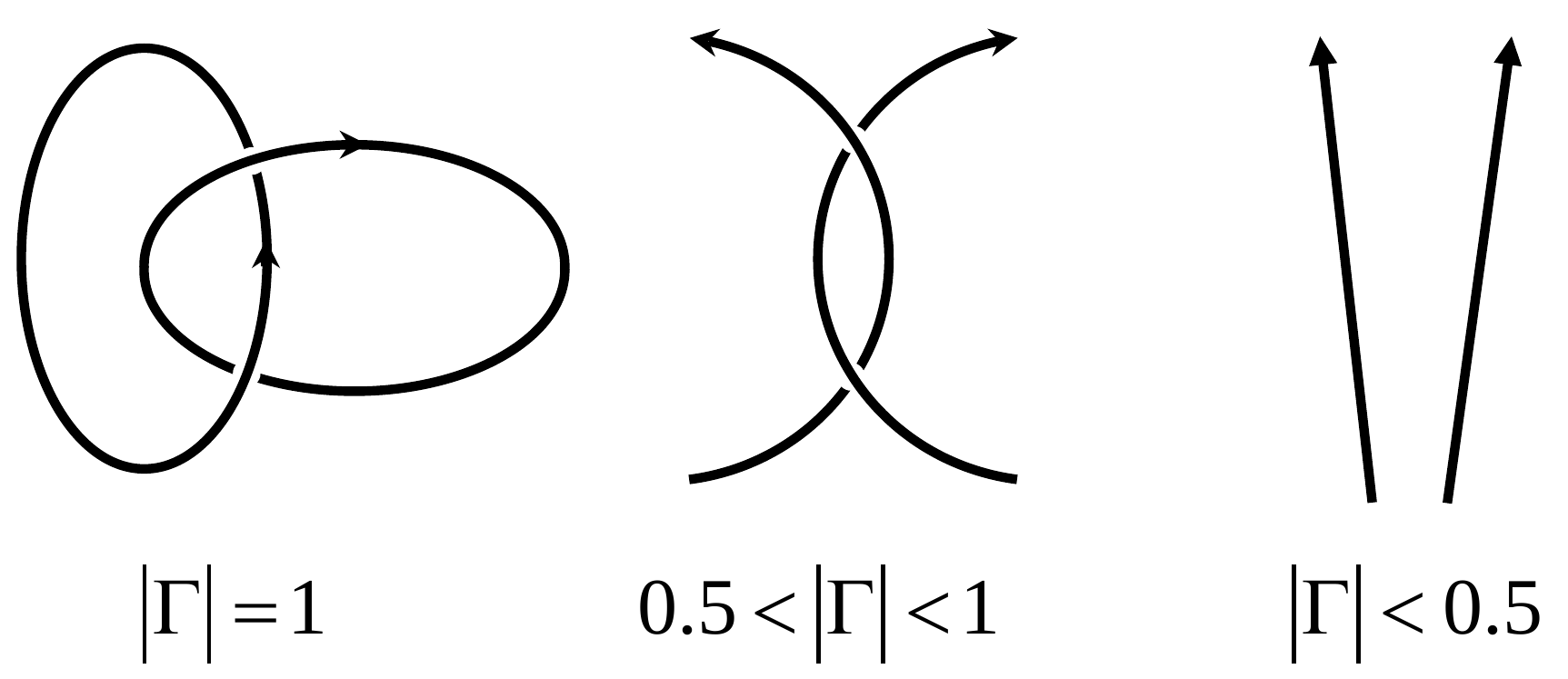}
\caption{Linking value of two curves for various configurations.}
\label{fig:writhe}
\end{figure}

%While the integral in Eq. \eqref{eq:gli} is interesting for closed curves, the Writhe matrix can be computed for open curves.
Intuitively, Eq. \eqref{eq:gli} counts how many windings around the first curve are completed and undone when traveling along the other curve as seen in Fig.~\ref{fig:writhe}. The entries $W_{i, j}$ of the Writhe matrix describe in which way the two line segments $s_1^i$ and $s_2^j$ pass each other. The absolute value of $W_{i, j}$ increases when the segments twist more or get closer and changes sign if the orientation of one segment is swapped.

%For two curves winding together, we need to evaluate how much they are twisting around each other. Writhe number \cite{fuller1971writhing} is a topological invariant that describe the linking of two directed curves in 3-d space.
%\todo{Lines, line segments, or curves? In 2d or 3d? Is the Writhe the same as the absolute value of the linking?  \qquad done}

%Since
%Intuitively, it represents how many turns one line wraps around the other, as is shown in Fig. \ref{fig:writhe}.

%This integral is also well-defined for piecewise linear curves. Considering we have two piecewise curves, $\gamma_1$ with $n_1$ segments and $\gamma_2$ with $n_2$ segments, if we compute the writhe number between the $i$-th segment $s^i_1$ on $\gamma_1$ and the $j$-th segment $s^j_2$ on $\gamma_2$, we can get a $n_1 \times n_2$ writhe matrix $W$ \cite{ho2010controlling}:
%
%\begin{equation}
%\begin{aligned}
%  W_{i,j} = w(s^i_1, s^j_2) \quad
%  for \ i &= 1,2,...n_1 \\
%  j &= 1,2,...n_2
%\end{aligned}
%\end{equation}
%
%where $W_{i,j}$ is the $(i,j)$-th element in the matrix.

%The writhe matrix calculates the topological relationship between each segment in two curves. Rather than only computing how much the two curves wrap each other, it can reflect how they wind, e.g. where is tight and where is loose, wraping from which direction, as is shown in Fig.~\ref{fig:heat}.
%\todo{Is the statement about seeing how tight it is true? I think the GLI of two rings is the same independent of the size of the rings.}
%From the heat maps we can see clearly where the value is high, which means a tight winding. Thus it is a good state representation in topological space.

%\todo{I think this is an effect of computing the GLI for segments of curves. If the two segments are close to each other, they cover a large section of angles. This can clearly be seen in the viewpoint covering interpretation (I'll explain it in meeting, there is a picture in my thesis too).}

\subsection{Laplacian Coordinates}
\label{sec:laplace}

Laplacian coordinates \cite{chung1997spectral, zhou2005large} describe the spacial relationship of points $p \in \mathbb{R}^n$ that are vertices of a graph $G =(V, E)$ relative to their neighborhood points $N_G(p) \subseteq V$ in the graph. These coordinates can describe local deformation of the graph but do not represent the relationship of indirectly connected vertices. The Laplacian coordinate $\delta_i$ for a point $p_i \in V$ is computed by a weighted sum of the neighborhood points,
\begin{equation}
\delta_i
=
p_i
-
\sum_{p_j \in N_G(p_i)} \alpha_{ij} p_j
,
\end{equation}
where $\alpha_{ij}$ is the normalization weight,
\begin{equation}
\alpha_{ij}=\frac{1}{|p_j-p_i| \,\sum_{p_k\in N_G(p_i)} | p_k - p_i |^{-1}}
,
\end{equation}
which sums up to 1 for each point $p_i$ so that this representation is invariant to scale \cite{zarubin2012hierarchical}.

\subsection{Representing the Robot-Humanoid Relationship}
\label{sec:representation}

\begin{figure}[]
\centering
\includegraphics[width=0.9\columnwidth]{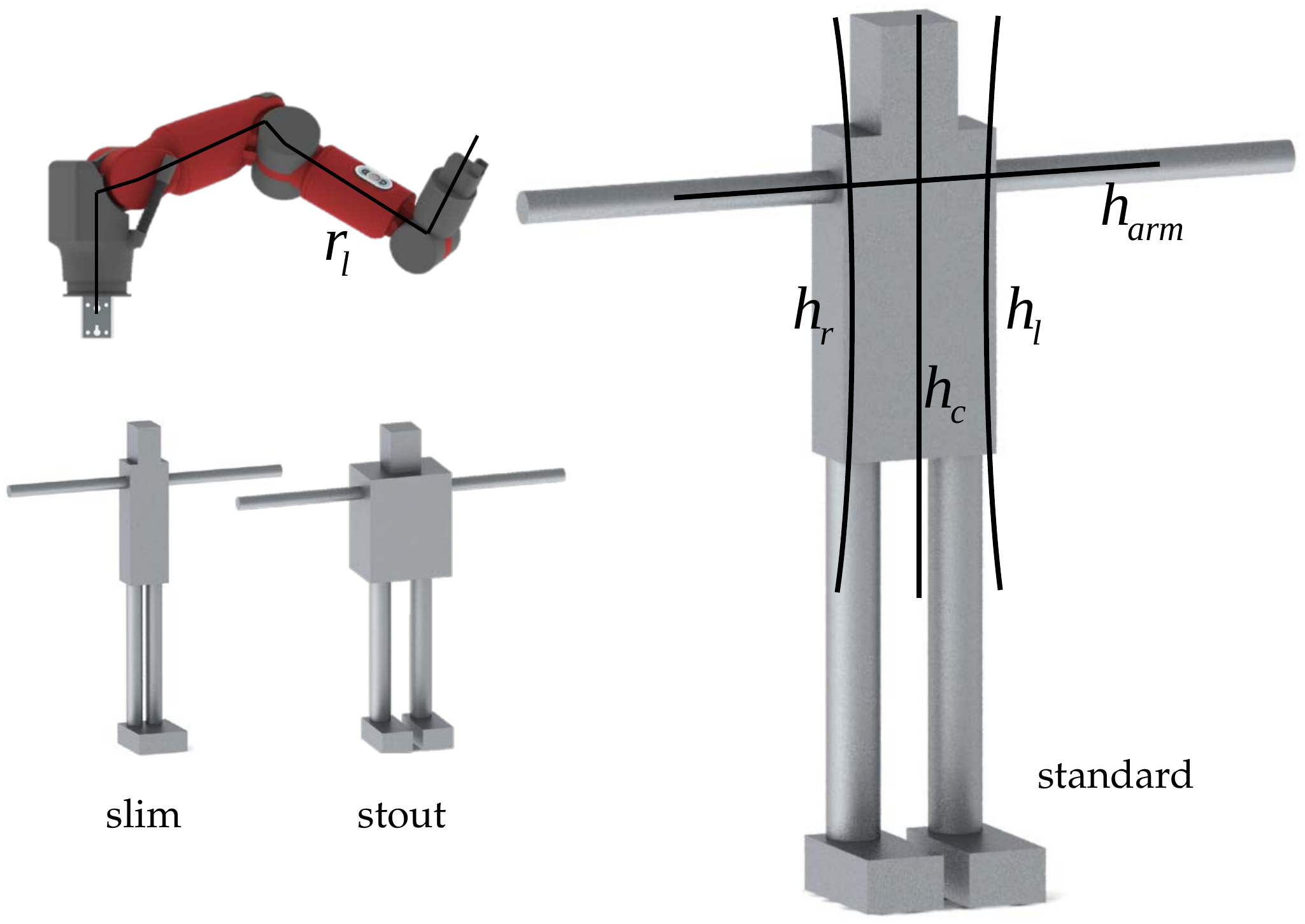}
\caption{The bodies of robot and humanoid are abstracted to curves. Each robot arm is represented by 7 line segments, and every curve in the humanoid is represented by 10 line segments. We train the policy with the standard model but also test the policy with the slim and stout models.}
\label{fig:humanoid}
\end{figure}

\begin{figure}[]
\centering
\includegraphics[width=0.4\columnwidth,trim={0cm 1cm 0cm 0cm}, clip]{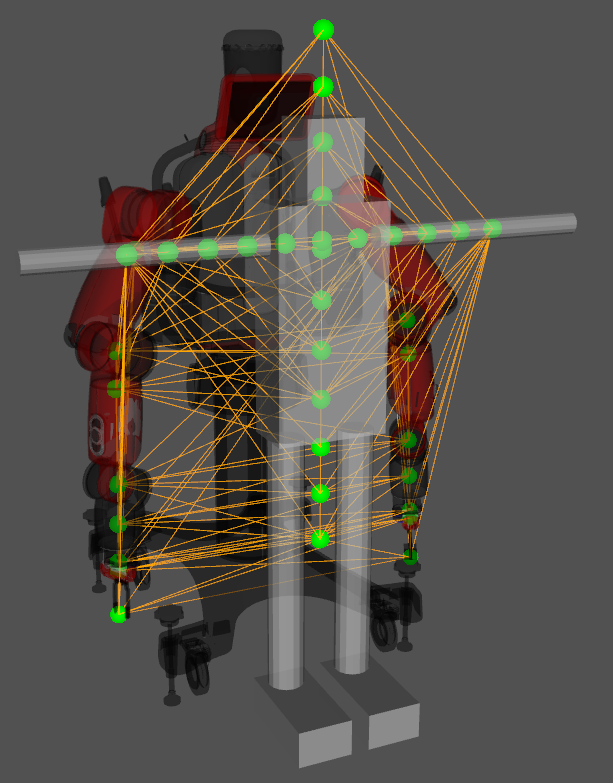}
\caption{For Laplacian coordinates we construct the graph $G$ from points on the curves in robot's arms and the humanoid's body. The orange lines are edges connecting the green vertices.}
\label{fig:graph}
\end{figure}

For our two motion generation scenarios, we combine Writhe matrix and Laplacian coordinates to represent the robot-humanoid relationship, similar to \cite{ivan2013topology, ho2010spatial}. To this end, we abstract the bodies of the robot and the humanoid into a set of curves consisting of line segments, as seen in Fig.~\ref{fig:humanoid}. This has the advantage that non-essential features of the bodies' geometry can be ignored by the learning algorithm.

\paragraph*{\textbf{For the robot}} We are only interested in the robot's arms and ignore all other body parts because only the arms should be used in the interaction. We introduce one curve for the right arm and one curve for the left arm, $r_{\mathrm{r}}$ and $r_{\mathrm{l}}$. Each curve has 7 line segments. The curves run from the base of the arms through the center of the links to the end of the arms where the tool can be attached.

\paragraph*{\textbf{For the humanoid}} We want to consider interaction with the arms and the torso. Therefore, we introduce one curve for the arms, $h_{\mathrm{arm}}$, one curve through the neck and the center of the torso, $h_{\mathrm{c}}$, and two curves each running though the shoulder and side of the torso, $h_{\mathrm{r}}$ and $h_{\mathrm{l}}$. Each of the four curves has 10 line segments. The curves $h_{\mathrm{c}}$, $h_{\mathrm{r}}$ and $h_{\mathrm{l}}$ are slightly longer than the torso and the curve $h_{\mathrm{arm}}$ ends approximately at the humanoid's elbows. For convenience we use superscript notation to refer to the upper and lower half of the curves in the torso as $h_{\mathrm{r}}^{\mathrm{upper}}$ and $h_{\mathrm{r}}^{\mathrm{lower}}$.

\begin{figure}[]
\centering
  \begin{subfigure}[b]{0.35\columnwidth}
    \includegraphics[width=\columnwidth]{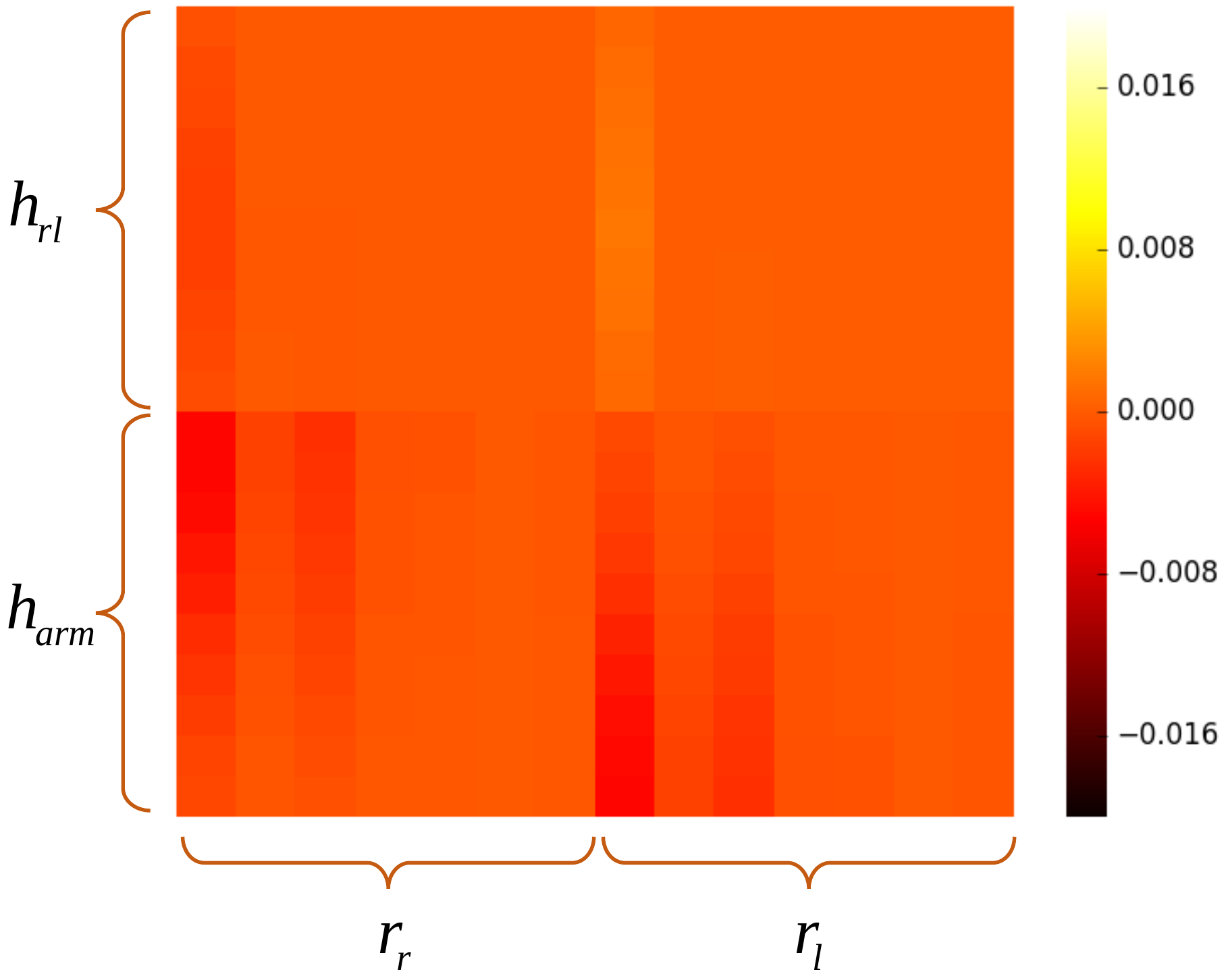}
    \caption{Initial $W_{\mathrm{U}}$}
    \label{fig:heat_writhe0}
  \end{subfigure}
  \begin{subfigure}[b]{0.35\columnwidth}
    \includegraphics[width=\columnwidth]{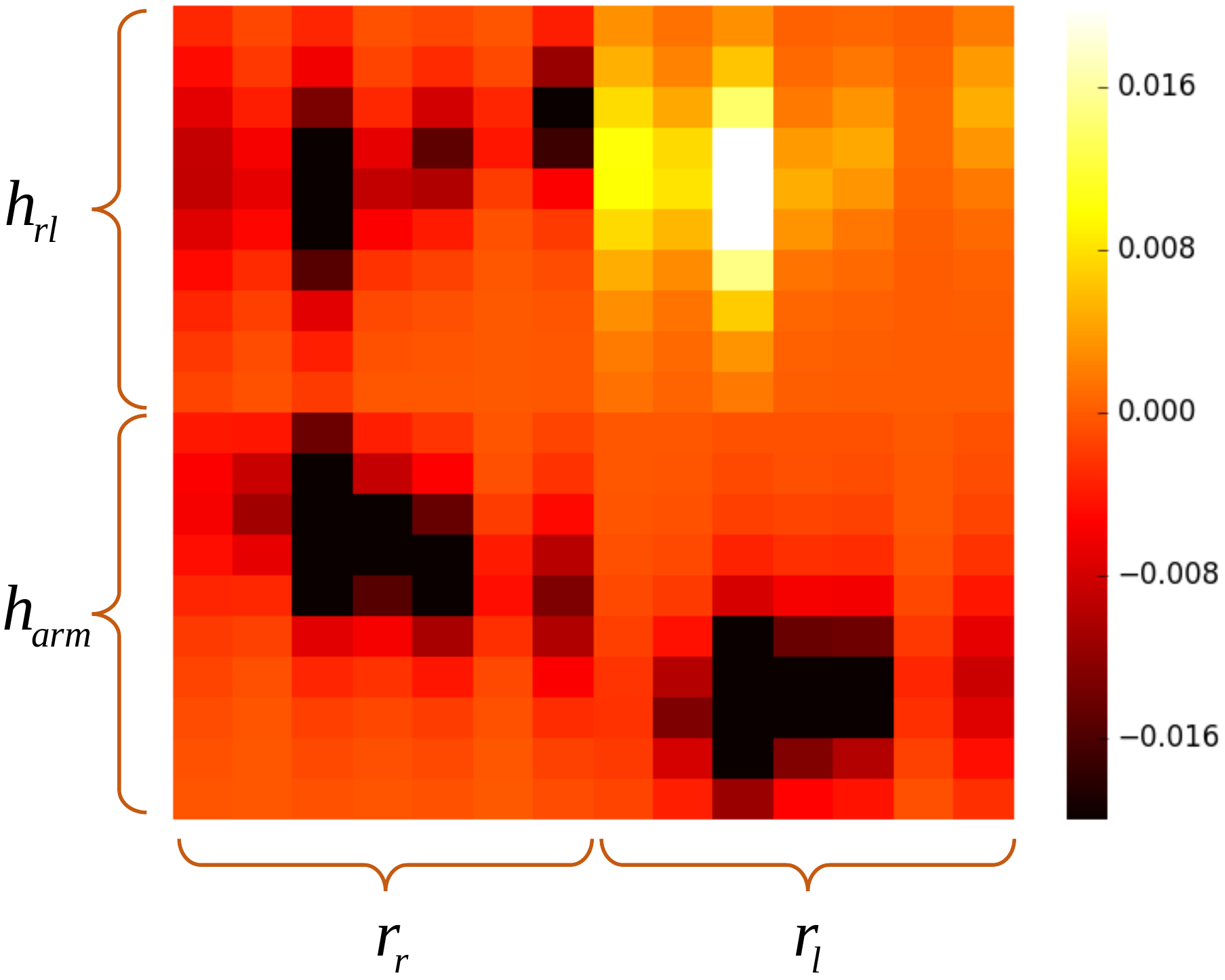}
    \caption{Final $W_{\mathrm{U}}$}
    \label{fig:heat_writhe22}
  \end{subfigure}
  \begin{subfigure}[b]{0.35\columnwidth}
    \includegraphics[width=\columnwidth]{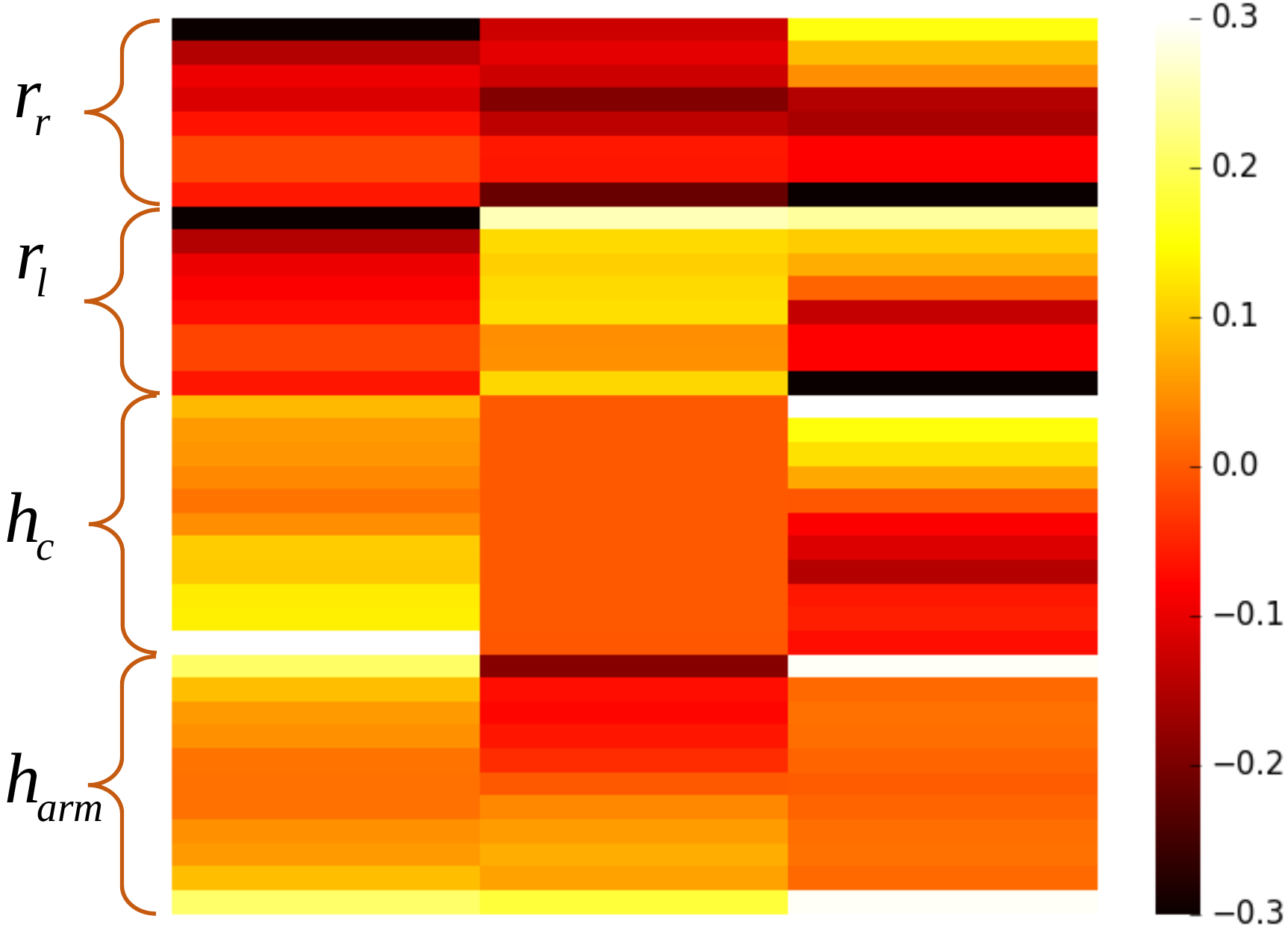}
    \caption{Initial $L_{\mathrm{U}}$}
    \label{fig:heat_lap0}
  \end{subfigure}
  \begin{subfigure}[b]{0.35\columnwidth}
    \includegraphics[width=\columnwidth]{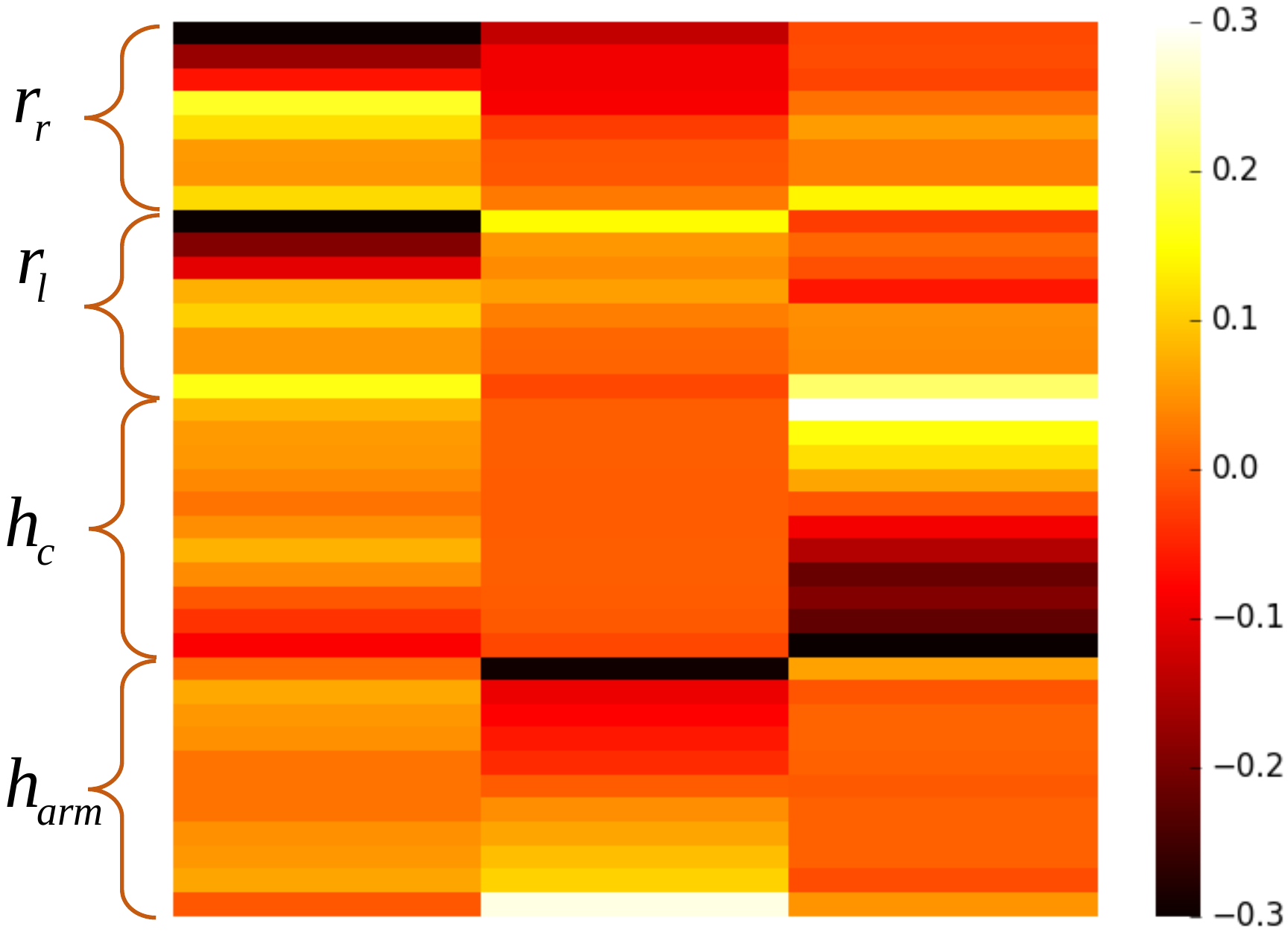}
    \caption{Final $L_{\mathrm{U}}$}
    \label{fig:heat_lap22}
  \end{subfigure}
\caption{The Writhe matrix $W_{\mathrm{U}}$ and the Laplacian coordinates $L_{\mathrm{U}}$ in initial state and final state for the upright scenario. The increase in linking between different body parts is clearly seen in $W_{\mathrm{U}}$.}
\label{fig:heat}
\end{figure}

Based on the curves defined above, we define two representations for interaction with the humanoid. One for the case where the humanoid is \emph{upright} (see Fig. \ref{fig:robot}(a)) and one where the humanoid is \emph{horizontal} (see Fig. \ref{fig:robot}(b)) in front of the robot. The two cases require different behaviors and using different representations allows better modeling of the relevant relationships. Below, we use the notation $W(\gamma_1, \gamma_2)$ for the Writhe matrix of curves $\gamma_1$ and $\gamma_2$.

\paragraph*{\textbf{Upright Pose}}
We define one combined Writhe matrix, $W_{\mathrm{U}} \in \mathbb{R}^{20 \times 14}$, from the robot's and humanoid's curves,
\begin{equation}
W_{\mathrm{U}}
=
\begin{pmatrix}
W(h_{\mathrm{r}}, r_{\mathrm{r}})
&
W(h_{\mathrm{l}}, r_{\mathrm{l}})
\\
W(h_{\mathrm{arm}}, r_{\mathrm{r}})
&
W(h_{\mathrm{arm}}, r_{\mathrm{l}})
\end{pmatrix}
.
\end{equation}
This captures the winding relationship between the robot's arms and the closest side of the humanoid's torso as well as the humanoid's arms. The matrix $W_{\mathrm{U}}$ is visualized in Fig. \ref{fig:heat}(a)\&(b).
For the matrix of Laplacian coordinates $L_{\mathrm{U}} \in \mathbb{R}^{38 \times 3}$ which captures the spacial relative distance relationship, we define the graph $G = (V, E)$ with 16 vertices from $r_{\mathrm{r}}$ and $r_{\mathrm{l}}$ and 22 vertices from $h_{\mathrm{c}}$ and $h_{\mathrm{arm}}$. The edges $E$ are defined by Delaunay triangulation of $V$ \cite{si2005meshing}. The graph $G$ is illustrated in Fig.~\ref{fig:graph} and $L_{\mathrm{U}}$ is visualized in Fig. \ref{fig:heat}(c)\&(d).

\paragraph*{\textbf{Horizontal Pose}}
We define $W_{\mathrm{H}} \in \mathbb{R}^{15 \times 14}$ from the robot's arm curves and the humanoid's torso curves,
\begin{equation}
W_{\mathrm{H}}
=
\begin{pmatrix}
W(h_{\mathrm{r}}^{\mathrm{upper}}, r_{\mathrm{r}})
&
W(h_{\mathrm{r}}^{\mathrm{lower}}, r_{\mathrm{l}})
\\
W(h_{\mathrm{c}}^{\mathrm{upper}}, r_{\mathrm{r}})
&
W(h_{\mathrm{c}}^{\mathrm{lower}}, r_{\mathrm{l}})
\\
W(h_{\mathrm{l}}^{\mathrm{upper}}, r_{\mathrm{r}})
&
W(h_{\mathrm{l}}^{\mathrm{lower}}, r_{\mathrm{l}})
\end{pmatrix}
.
\end{equation}
This captures the winding relationship between the robot's arms and the upper and lower part of the humanoid's torso separately.
For the matrix $L_{\mathrm{H}} \in \mathbb{R}^{49 \times 3}$, we define the graph $G = (V, E)$ with 16 vertices from $r_{\mathrm{r}}$ and $r_{\mathrm{l}}$ and 33 vertices from $h_{\mathrm{c}}$, $h_{\mathrm{l}}$, and $h_{\mathrm{r}}$. The edges $E$ are again defined by Delaunay triangulation of $V$.

\section{LEARNING TO GENERATE MOTIONS}
\label{sec:policy_learning}

We assume that our robot is compliant and has a low-level controller that accepts desired joint angles and drives the robot's motors while monitoring force and effort limits. That means that our motion policy can command the robot joint angles without directly considering velocities, kinematic, or contacts and we can still get close interaction between the robot and the humanoid. Below we explain how we train the motion policy with deep reinforcement learning. For this, we first define a reinforcement learning problem and model the task with a reward function in Sec. \ref{sec:statement}. In Sec. \ref{sec:ppo} we give details about the reinforcement learning algorithm, and in Sec. \ref{sec:structure} we explain the network structure.

\subsection{Learning Problem}
\label{sec:statement}

For setting up a reinforcement learning problem to train the motion generation policy, we need to define a state space $S$, an action space $A$, and a reward function $r_t$ for each time step $t$. Below, we first describe the motion that we want to generate in the two interaction scenarios introduced in Sec.~\ref{sec:representation}, and then formulate the reinforcement learning problem used to learn the policies.

\paragraph*{\textbf{Upright Pose Scenario}}

The humanoid is positioned upright in front of the robot and we want to achieve a state in which the robot can lift and drag the humanoid backwards, such as in a \emph{shoulder drag}. To achieve this, we want the robot to move its arms forward and hold the humanoid tightly below the shoulders as seen in Fig. \ref{fig:robot}(a).

\paragraph*{\textbf{Horizontal Pose Scenario}}

The humanoid is positioned horizontally in front of the robot and we want to achieve a state in which the robot can lift and carry the humanoid, such as in a \emph{cradle lift carry}. This is achieved by moving the robot's arms forward and under the humanoid to hold the humanoid tightly from below as seen in Fig. \ref{fig:robot}(b).

\paragraph*{\textbf{Action Space and Control}}

The action space $A = \mathbb{R}^{14}$ is the same in both scenarios and consists of desired changes in joint angles. Therefore, the sum of an action $a \in A$ and the vector of current joint angle $J$ define a new target for the low-level controller, $J+a$. In every time step, the robot has 2 seconds to reach the desired joint angle $J+a$. After that or when the target is reached earlier, the next time step starts.

\paragraph*{\textbf{State Space}}

In both scenarios, we define the state space $S$ by a combination of the Writhe matrix and the Laplacian coordinates. For the upright case this combination has $20 \times 14 +  38 \times 3 = 394$ dimensions and for the horizontal case it has $15 \times 14 +  49 \times 3 = 357$ dimensions. This state space captures spacial relationships as well as local geometric properties.

\paragraph*{\textbf{Reward Function}}

For the reward function, we first define the total linking values $\Gamma_{\mathrm{U}}$ and $\Gamma_{\mathrm{H}}$ which sum up the absolute value of linking between the curves that are used to construct the combined Writhe matrices $W_{\mathrm{U}}$ and $W_{\mathrm{H}}$,
\begin{align}
\Gamma_{\mathrm{U}}
=&{}
| \Gamma(r_{\mathrm{l}}, h_{\mathrm{l}})| + | \Gamma(r_{\mathrm{l}}, h_{\mathrm{arm}})|+
\notag
\\
&{}
| \Gamma(r_{\mathrm{r}}, h_{\mathrm{r}})| + | \Gamma(r_{\mathrm{r}}, h_{\mathrm{arm}})|
 \label{eq:tlU}
\intertext{and}
\Gamma_{\mathrm{H}}
=&{}
| \Gamma(r_{\mathrm{r}}, h_{\mathrm{l}}^{\mathrm{upper}})| +
| \Gamma(r_{\mathrm{r}}, h_{\mathrm{c}}^{\mathrm{upper}})| +
| \Gamma(r_{\mathrm{r}}, h_{\mathrm{r}}^{\mathrm{upper}})| +
\notag
\\
&{}
| \Gamma(r_{\mathrm{l}}, h_{\mathrm{l}}^{\mathrm{lower}})| +
| \Gamma(r_{\mathrm{l}}, h_{\mathrm{c}}^{\mathrm{lower}})| +
 | \Gamma(r_{\mathrm{l}}, h_{\mathrm{r}}^{\mathrm{lower}})|
 \label{eq:tlH}
 .
\end{align}

The total linking values in Eq. \eqref{eq:tlU} and \eqref{eq:tlH} capture the global property of how much the involved curves wind around each other. We select the curves precisely so that these values are maximized in robot-humanoid configurations that are required in our two scenarios. Therefore, we define the reward in terms of total linking value, its recent increment and a punishment term:
\begin{equation}
\begin{aligned}
r_t = \,
&\beta_1 \,
\left(
10 \, \Delta_t
+
(\Gamma_{\mathrm{U}})_t - \Gamma_{\mathrm{ref}} \right)
-
\\
&\beta_2 \, ( \max(0,z_{\mathrm{r}}) + \max(0,z_{\mathrm{l}}))
.
\end{aligned}
\label{eq:reward}
\end{equation}
where $\beta_1, \beta_2$ are scale factors, $\Gamma_{\mathrm{ref}}$ is an offset value, and $\Delta_t = (\Gamma_{\mathrm{U}})_t - (\Gamma_{\mathrm{U}})_{t-1}$ is the last increment in total linking. The second line considers the mean height difference between the robots left and right arm and the humanoid shoulders, $z_{\mathrm{l}}$ and $z_{\mathrm{r}}$, and is 0 when the arms are below the shoulders. This makes sure that the robot holds from below and can actually lift or carry the humanoid with its arms. Eq. \eqref{eq:reward} is defined analogously for the horizontal scenario.

\subsection{Reinforcement Learning}
\label{sec:ppo}

The policy is trained with Proximal Policy Optimization (PPO) \cite{schulman2017proximal}, which is an actor-critic reinforcement learning method. Actor-critic methods maintain both, a policy estimate (the actor) $\pi( a | s; \theta^{\pi})$, which maps the states to actions, and a value estimate (the critic) $V( s; \theta^{V})$, which predicts the discounted sum of future rewards. Both are modeled as neural networks with their respective parameters $\theta^{\pi}$ and $\theta^{V}$.

During learning, the critic's loss, $\mathcal{L}_V(\theta^V)$, minimizes the difference between actual return $R_t = \sum_{i=t}^{\infty}\gamma^{i-t}r_i$ and estimated value, where $\gamma$ is the discount factor.
%
%The optimization of the value network is to reduce the loss between the return $R_t = \sum_{i=t}^{\infty}\gamma^{i-t}r_i$ and the value estimator
%
%\begin{equation}
%  \mathcal{L}_V(\theta^V)=\mathbb{E}\left[\left(R_t-V(s_t;\theta^V)\right)^2\right]
%\end{equation}
%
%where $\gamma$ is the discount factor.
%
The actor's objective $\mathcal{J}_{\mathrm{ppo}}(\theta^\pi)$ maximizes the advantage function, which estimates the difference between the value of output action and all actions.
%
%The optimization of the policy network is to directly compute the estimator of the policy gradient as
%
%\begin{equation}
%  \triangledown_{\theta^{\pi}}\mathcal{J}(\theta^\pi)=\mathbb{E}[\triangledown_{\theta^\pi}\log\pi(a_t|s_t;\theta^\pi)\hat{A}_t]
%\end{equation}
%
%where $\mathcal{J}$ is the objective, $\pi$ is stochastic policy, $\theta^{\pi}$ is the policy parameters, $\hat{A}_t$ is the estimator of the advantage function at timestep $t$. For a rollout with time steps $\{1,...,t,...T\}$, it is computed as
%
%\begin{equation}
%\hat{A}_t=\sum_{i=0}^{T-t-1}\gamma^ir_{t+i}+\gamma^{T-t}V(s_{T};\theta^V)-V(s_t;\theta^V)
%\end{equation}
%
%To be more efficient and robust, PPO utilizes minibatch update on a new objective
%
%\begin{equation}
%\mathcal{J}_{ppo}(\theta^\pi)=
%\mathbb{E}\left[min\left\{
%r_t\hat{A}_t,\
%clip(r_t,1-\epsilon,1+\epsilon)\hat{A}_t
%\right\}\right]
%\end{equation}
%
%where $\epsilon$ is the clipping factor and usually set $0.2$, $r_t$ is the probability ratio between new policy and the policy before the update,
%
%\begin{equation}
%r_t=\frac{\pi(a_t|s_t;\theta_\pi)}{\pi_{old}(a_t|s_t;\theta_\pi)}
%\end{equation}
%
To encourage exploration \cite{mnih2016asynchronous}, we add the entropy of the policy, $E(\pi)$, such that the final loss is given by
\begin{equation}
\mathcal{L}=c_1\mathcal{L}_V(\theta^V)-\mathcal{J}_{\mathrm{ppo}}(\theta^\pi)-c_2E(\pi(s_t;\theta^\pi))
\end{equation}
where $c_1$ is the value loss coefficient and $c_2$ is the entropy regularization coefficient.

% \todo{Read and check this section because it has been extremely simplified. We need to say something more about the actor's loss. I did not understand what was written there. It was about that it is not the same thing as policy gradient?}

\subsection{Network Architecture}
\label{sec:structure}

\begin{figure}[]
\centering
\includegraphics[width=0.85\columnwidth]{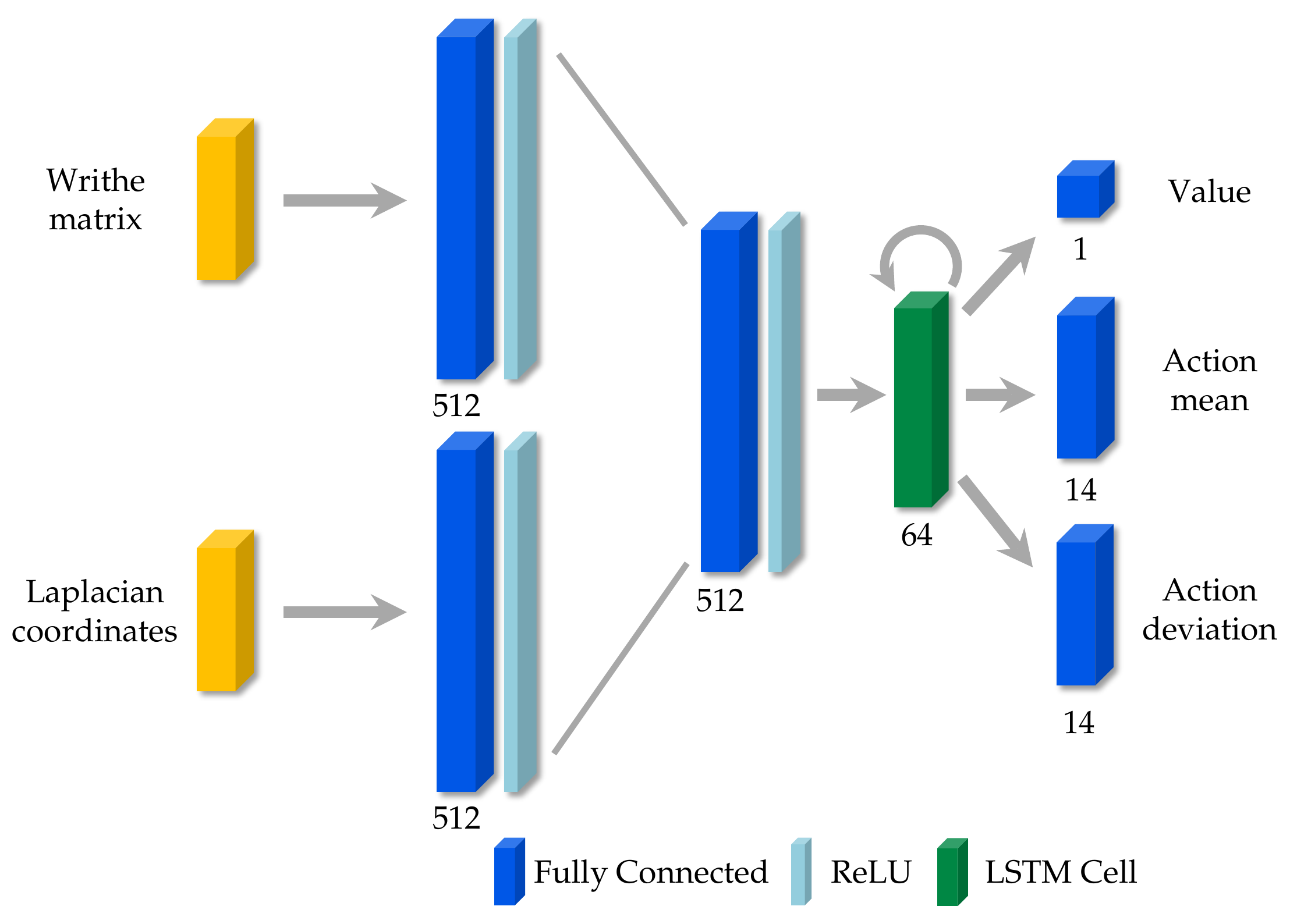}
\caption{Network structure: The network is composed of a multi-layer perceptron (MLP) base, LSTM unit, and output heads. The inputs are Writhe matrix and Laplacian coordinates and outputs are the scalar value (the critic) and the action vector with its standard deviation (the actor). The fully-connected layers in the base have ReLU activation while action mean and deviation are Tanh and Softplus, respectively.}
\label{fig:network}
\end{figure}

For reinforcement learning with PPO, we define an actor-critic network as shown in Fig.~\ref{fig:network}. We feed the Writhe information and the Laplacian coordinates into separate first layers. First two layers extract useful features from the state vector which are then fed into a recurrent neural network. The Long Short-Term Memory (LSTM) \cite{hochreiter1997long} unit allows the model to remember previous states. Using three independent layers, the LSTM state is then mapped to the value estimate, the action mean and the action variance, where the last two define the probabilistic policy $\pi$ as a multivariate Gaussian.

%through the value head layer and mapped to an action vector $\mathbf{a}_{\text{mean}}$ plus its standard deviation $\boldsymbol{\sigma}$ through the action head layer. Then the action of the robot is sampled from the normal distribution as
%
%\begin{equation}
%\mathbf{a}
%=
%\mathit{Normal}(\mathbf{a}_{\text{mean}}, \boldsymbol{\sigma})
%\end{equation}
%
%where $\mathit{Normal}()$ is a normal distribution sampling function.

%There are two states input to the network, one is writhe matrix $W$, containing the winding information, and the other is Laplacian coordinates $L$, containing the information of the relative spatial distance. They are fed forward into two independent layers and then concatenated together and forwarded to next layer.

%\todo{Does that mean that the writhe matrix is not considered as a matrix and there are no convolutions?
%\qquad yes}

%Since running this task in one episode is a time serious process, it should be important to remember what the robot has done in this episode.

%\todo{Does that mean that in fact the state representation (the thing that is feed to the policy network) is actually the LSTM state? This would mean that the representation you construct is not the state but the observation.

%Also, do you use LSTM because the representation you construct is not Markovian?

%solved}

%% file: includes/experiments.tex
% !TEX root =  ../main.tex

\begin{figure*}[]
\centering
  \begin{subfigure}[b]{0.31\columnwidth}
    \includegraphics[height=3.5cm, trim={0cm 0cm 0cm 0cm}, clip]{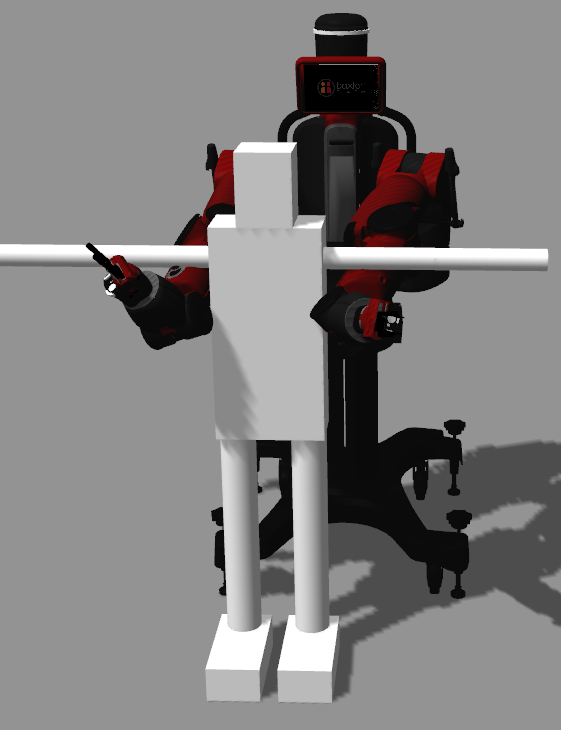}
    \caption{$\Gamma_{\mathrm{U}}=1.5$}
    \label{fig:w=15}
  \end{subfigure}
  \begin{subfigure}[b]{0.31\columnwidth}
    \includegraphics[height=3.5cm, trim={0cm 0cm 0cm 0cm}, clip]{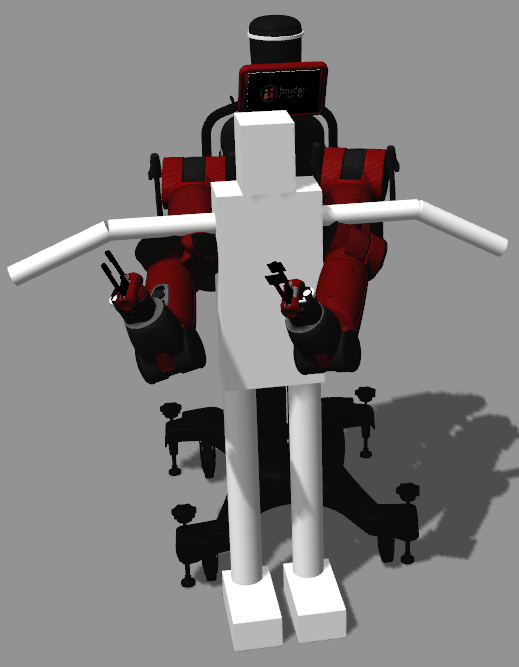}
    \caption{$\Gamma_{\mathrm{U}}=1.7$}
    \label{fig:w=17}
  \end{subfigure}
  \begin{subfigure}[b]{0.31\columnwidth}
    \includegraphics[height=3.5cm, trim={0.5cm 0cm 0.5cm 0cm}, clip]{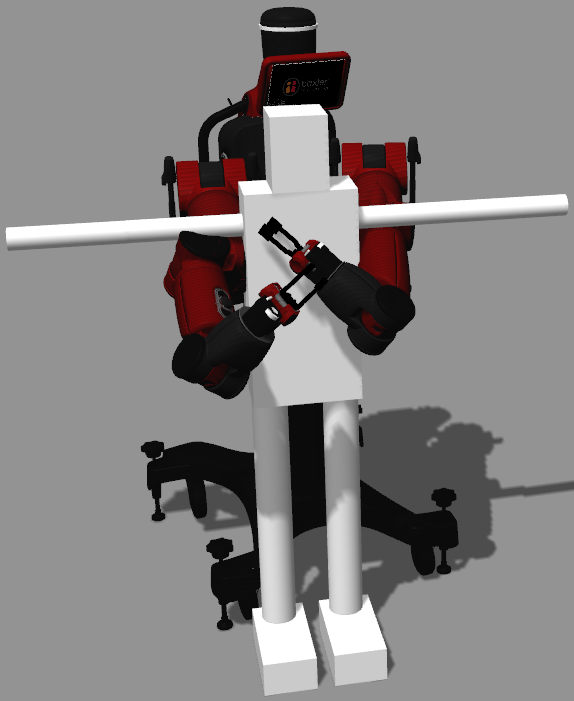}
    \caption{$\Gamma_{\mathrm{U}}=1.9$}
    \label{fig:w=19}
  \end{subfigure}
  \begin{subfigure}[b]{0.3\columnwidth}
    \includegraphics[height=3.5cm, trim={0cm 0cm 0cm 0cm}, clip]{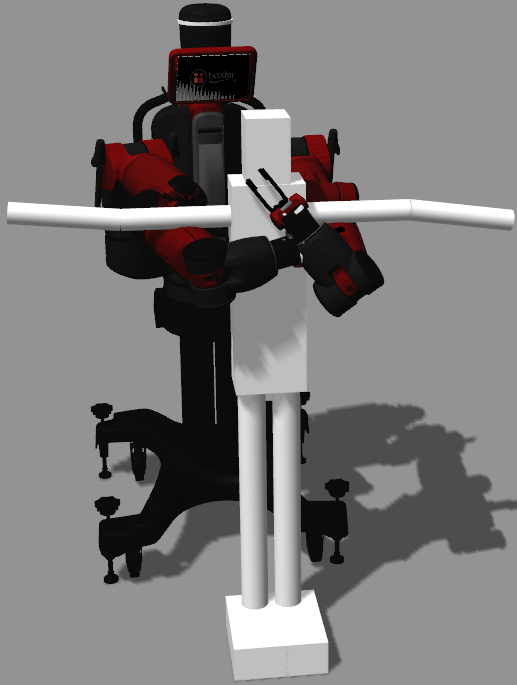}
    \caption{$\Gamma_{\mathrm{U}}=2.2$}
    \label{fig:slim}
  \end{subfigure}
  \begin{subfigure}[b]{0.3\columnwidth}
    \includegraphics[height=3.5cm, trim={1cm 0cm 1cm 0cm}, clip]{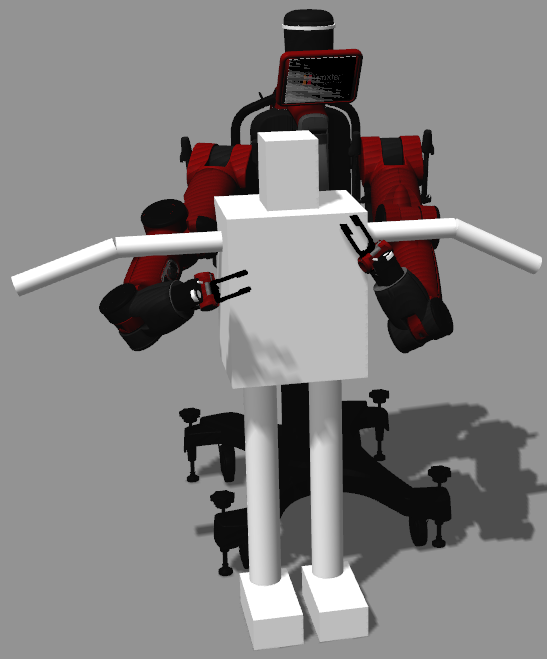}
    \caption{$\Gamma_{\mathrm{U}}=1.6$}
    \label{fig:fat}
  \end{subfigure}
  \begin{subfigure}[b]{0.45\columnwidth}
    \includegraphics[height=3.5cm, trim={0.5cm 0cm 0.5cm 0cm}, clip]{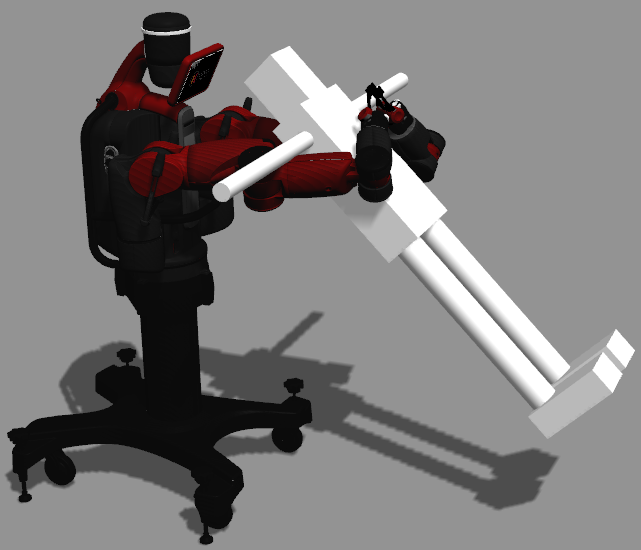}
    \caption{$\Gamma_{\mathrm{U}}=2.1$}
    \label{fig:floating}
  \end{subfigure}

\caption{Example holding actions executed by Baxter robot in different scenarios. (a-c) Example holdings on the standard humanoid model with the reference linking number $\Gamma_{\mathrm{ref}}=1.5$ and linking number $1.7, 1.9$. (d-e) Examples showing holding actions on different humanoid models not involved in training. (f) A holding action applied on a humanoid model floating in water in a non-upright pose. More executions can be found in https://youtu.be/Al-QZl-WGlw.}
\label{fig:holding}
\end{figure*}

\section{EXPERIMENTS}
\label{sec:experiments}

We evaluate our work from $3$ perspectives: \emph{1)} we describe the observations in our training process and analyze the network's performance based on the employed topological and spatial representations, in comparison to using a simple position representation; \emph{2)} we quantitatively evaluate the trained policy in terms of the scale of target humanoid model and simulated perception uncertainty; \emph{3)} we present qualitative experiments of example application scenarios of the proposed WAM as well as demonstrating a real world example.

The experiments were conducted in Gazebo with a Baxter robot and differently scaled humanoid models and focused on the upright humanoid. In both training and evaluation, we simulate dynamic humanoid models by oscillating the model's velocity in the vertical direction according to a sinusoidal function with peak-to-peak distance of $25$ cm. For every episode, the humanoid's model is always initialized with its back facing the robot and we randomize its position within a $40\times40$ cm$^2$ squared region in front of the robot. The step limit $T_{\text{max}}$ for each episode is set to 10 so the total time for one episode is within 20s. Most time is spent on robot movement while the network forward time is only about $0.8$ millisecond.

% All the training and evaluations were run on a machine with a Intel Core i7 6850K @ 3.6Ghz and a Nvidia GeForce 1080 Ti GPU.

\subsection{Network Training}

For training the network, we set the parameters as listed in Table~\ref{tab:parameters}, and used only the standard humanoid model in Fig.~\ref{fig:humanoid}. For choosing the $\Gamma_{\mathrm{ref}}$ value, we empirically checked a range of reference linking values and decided to set it as $\Gamma_{\mathrm{ref}}=1.5$. As shown in Fig.~\ref{fig:holding}(a), when the total linking number is $1.5$, the robot arms start to form a holding around the humanoid model. In the process of training, we updated the network $4$ times after each episode using the Adam optimizer \cite{kingma2014adam} based on the last $4$ experience batches.

\begin{table}[!t]
\centering
\caption{Learning Parameters}
\begin{tabular}{r c c}
\toprule
Parameter & Notation & Value \\
\midrule
Episode limit & $T_{\text{max}}$ & 10\\
\rowcolor[gray]{0.95}
Reward scale factor & $\beta_1, \beta_2$ & $5,1$\\
Reference Linking & $\Gamma_{\mathrm{ref}}$ & 1.5 \\
\rowcolor[gray]{0.95}
Learning rate & $\eta$ & $10^{-4}$ \\
Discount factor & $\gamma$ & $0.99$ \\
\rowcolor[gray]{0.95}
Value loss coefficient & $c_1$ & $0.5$ \\
Entropy regularization coefficient & $c_2$ & $0.01$ \\
\bottomrule
\end{tabular}
\label{tab:parameters}
\end{table}

\begin{figure}[htb]
\centering
\input{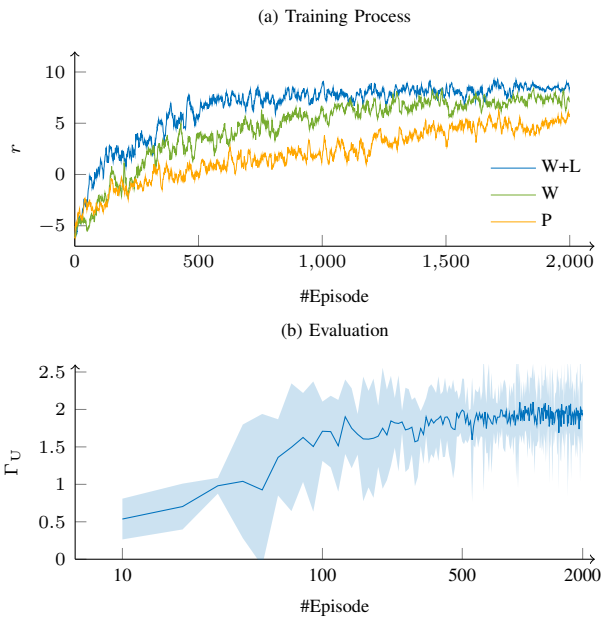}
\input{images/evalu_writheinter.tex}
\caption{Training results: (a) The reward mean $r$ of every episode is shown as the training progresses. To filter out the reward noise, each plotted curve is smoothed locally using the $10$ neighboring points. W: Writhe matrix, L: Laplacian coordinates and P: Landmark positions (b) The total linking number $\Gamma_{\mathrm{U}}$ achieved by the configuration (W+L) is evaluated online during training. The number of episode is plotted in log-scale. }
\label{fig:exp}
\end{figure}

In order to evaluate the effectiveness of the proposed representations, we trained the network using $3$ different input spaces: i) as shown in Fig.~\ref{fig:network}, a network is trained using both the Writhe matrix and the Laplacian coordinates; ii) a network is trained with only the Writhe matrix as the input; and iii) without using the representations developed in this work, we directly use a $3 \times 38$ matrix, which contains $38$ position coordinates of the landmark points shown in Fig.~\ref{fig:graph}, as the input to the network for comparison. We repeated the training for each of the $3$ cases for $5$ times and report the average results in Fig.~\ref{fig:exp}.

As seen in Fig.~\ref{fig:exp}(a), when using both the Writhe matrix and Laplacian coordinates, the network was able to converge after experiencing about $600$ episodes and achieved the best result over the $3$ test cases. During the training, to see the performance without exploration deviation noise, we also conducted online evaluation for case i) as shown in Fig.~\ref{fig:exp}(b). For this, we ran the trained policy without variance after every $10$ episodes to try to hold a dynamic humanoid model using $10$ actions, and recorded the resulted total linking number $\Gamma_{\mathrm{U}}$. The result indicates that the network has finally learned how to do this task with a high $\Gamma_{\mathrm{U}}$ value of around $2$. Comparing to the reference $\Gamma_{\mathrm{ref}}=1.5$ and as exemplified in Fig.~\ref{fig:holding}, this will provide robust behaviors to hold the humanoid model.

% As is shown in the results and observed in the experiments, without writhe matrix the network cannot solve this task. The robot does not understand this task and moves casually. With only the writhe matrix can handle it basically but the performance is not stable. Sometimes the robot makes some mistakes, the reason of which is maybe the information in the writhe matrix is not enough to represent the state completely. This proves that the Laplacian coordinates provides more information about the spacial distance relationship and is helpful to this task.
% \todo{More or additional?}

In comparison to case i), using only the Writhe matrix to train the network performed worse after the training was converged after $1500$ episodes. This is because of two reasons: firstly, the Writhe matrix by itself does not encode enough relative spatial information between the robot and the humanoid, it is not able to describe geometric interactions. More importantly, by definition different robot states can potentially result in the same Writhe matrix, which can confuse the network in many cases. Lastly, we can see that using only position information of landmark points performed the worst. In our evaluation, it was not able to execute the task even after convergence. This further implies the importance of using the topological representation, which essentially captures the winding interaction between links.

% And the results of the training with position state shown in Fig.~\ref{fig:exp}(b) reveals this problem cannot be well solved by classical 3-d coordinates input. Its performance is much lower than the topological representations.

% The comparison between the writhe matrix and Laplacian coordinates corresponding to a good holding and initial state of one episode is presented in Fig.~\ref{fig:heat}. From the comparison we can see at the beginning there is nearly no winding between the robot arms and the humanoid. After some actions of the robot, a holding state with high-value writhe matrix appears. The windings between the robot left arm $r_l$ and the humanoid torso $h_l$, robot left arm $r_l$ and humanoid arm $h_{arm}$, robot right arm $r_r$ and humanoid torso $h_r$, robot arm $r_r$ and humanoid arm $h_{arm}$ are displayed clearly in the writhe matrix.
%
% To evaluate the performance without exploration noise, we do online evaluation every 10 episodes. We remove the action deviation and give the action output of the network to the robot and test what writhe it can reach within 10 steps, as is shown in Fig.\ref{fig:exp} (c). The curve is averaged over 5 trials and its 95\% confidence interval is indicated as shadow. We can see that the performance rises steadily and $w_g$ reaches almost 2, which is in a good winding status.

\subsection{Novel Scenarios and Perception Uncertainty}
% After the training, we want to examine the performance and stability of the model. We define writhe $w_g$ over 1.5 counts a successful execution, and let robot execute this task within 10 steps. We ran the model trained after 2000 episodes for 5 trials, each with 100 episodes. The average success rate is 99\%, which indicates the performance of the robot is pretty stable.

% \todo{Is the a better way to show that the robot holds the person? The writhe value is not direct proof that the behavior is successful.}

Having trained the policy using only the standard humanoid model in Fig.~\ref{fig:humanoid}, we now evaluate its performance using differently shaped and scaled novel models. The trained policy has been applied on some novel humanoid models and a few examples are demonstrated in Fig.~\ref{fig:holding}(d-e). As we can observe, although the humanoid models possess relatively large differences in geometries, the trained policy guided the robot to move its arms around the torso and arms of the humanoid models, and was able to finally achieve holding actions with high linking numbers.

In addition, we quantitatively test the policy by applying it to the $3$ humanoid models in Fig.~\ref{fig:humanoid}. For each model, we randomize its initial position and keep it moving up and down in front of the robot within a $40 \times 40 $ cm$^2$ region for $100$ times $\times$ $5$ batches and let the network run for $10$ steps for each execution. An execution is successful if the final linking number $\Gamma_{\mathrm{U}}$ is greater than $1.5$. As reported in Table~\ref{tab:evaluation}, the policy performs well and achieves an average success rate of $99\%$ when evaluated with the standard model, which was adopted also in training. For the slim model, the policy performs equally well with a success rate of $98\%$. However, the performance drops to $92.6\%$ for the stout model. As one can observe, the stout model is shorter and wider, which is naturally more difficult to be wrapped around. Moreover, since the robot arms are kept away from each other by the wide torso, the maximum achievable linking number for this model is lower than the others limited by the length of the arms, it is therefore infeasible for the robot to achieve high linking number on it when the model is relatively far from the robot.

% To examine the generalization of the network model in topological space, we test the performance of the network model trained with original humanoid in the scenarios with novel different-shape humanoid models, a fat humanoid and a slim humanoid, as shown in Fig.~\ref{fig:humanoid}. The test setup is the same as the original humanoid. The success rate is still very high for the slim humanoid with average $98\%$ but lower for the fat humanoid with average $92.6\%$, as presented in TABLE~\ref{tab:evaluation} with the standard deviation. Intuitivly, the fat humanoid is shorter and wider, which is harder for the robot to hold with a high writhe number. It is limited by the length and the motion range of the robot arms and the holding with similar tightness gets lower score, as shown in Fig.~\ref{fig:holding}(e), where a good holding gets only 1.6. If we lower the bar, the robot can get 1.4 writhe with $99\%$ success rate.
%
% Besides, we put the humanoid floating on the air to test its generalization like Fig.~\ref{fig:holding}(f). The humanoid pose is fixed and its position is also random within a $40\times40$ region. The robot performs pretty well and finishes the task with $100\%$ success rate, which proves that the policy can handle various scenarios. Or, there is no much difference for these different scenes in the topological space. Thus, in fact not only the humanoids, any similar-shape objects are the same in topology space and can be held by the robot with this policy.

\begin{table}[t]
\centering
\caption{Success Rates}
\begin{tabular}{c c}
\toprule
Humanoid Model & Success Rate\\
\midrule
Standard & $99.00\%\pm1.10\%$ \\
\rowcolor[gray]{0.95}
Slim & $98.00\%\pm0.89\%$ \\
Stout & $92.60\%\pm1.96\%$ \\
% \rowcolor[gray]{0.95}
% Floating & $100.00\%\pm0.00\%$ \\
\bottomrule
\end{tabular}
\label{tab:evaluation}
\end{table}

% To test our policy's rubustness facing perception noise, we add Gaussian noise to the skeleton curves of the humanoid. We add the noise to every point in the curves and test how this influence the robot. The results presented in Fig.~\ref{fig:noisy} show the rubustness of the policy. With 0.1m noise there is nearly no performance reduction. Only with 0.2m even 0.5m noise the performance is declined.

\begin{figure}[t]
\centering
\input{images/test_noisy.tex}
\caption{Evaluation against perception noise $\sigma=0.1,0.2,0.3$m. The average linking number is plotted with its $95\%$ confidence interval for each action step.}
\label{fig:noisy}
\end{figure}
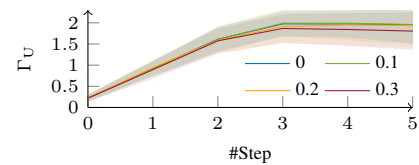

For evaluating the system robustness against the perception uncertainty, we simulate the perception errors for the landmark points using Gaussian distributions. In the presence of different magnitudes $\sigma$ of perception errors, we apply the trained policy on the standard humanoid model and recorded the achieved $\Gamma_{\mathrm{U}}$ against the movement step. This experiment is repeated for $100$ times for each $\sigma$ and the statistics is reported in Fig.~\ref{fig:noisy}. This result indicates that our trained policy is not significantly affected by the perception noise, since adopted topological representation is not sensitive to the absolute positions of landmark points.

\subsection{Qualitative Experiments}
\begin{figure}[]
\centering
  \begin{subfigure}[b]{0.37\columnwidth}
    \includegraphics[height=3.5cm, trim={0cm 0cm 0cm 0cm}, clip]{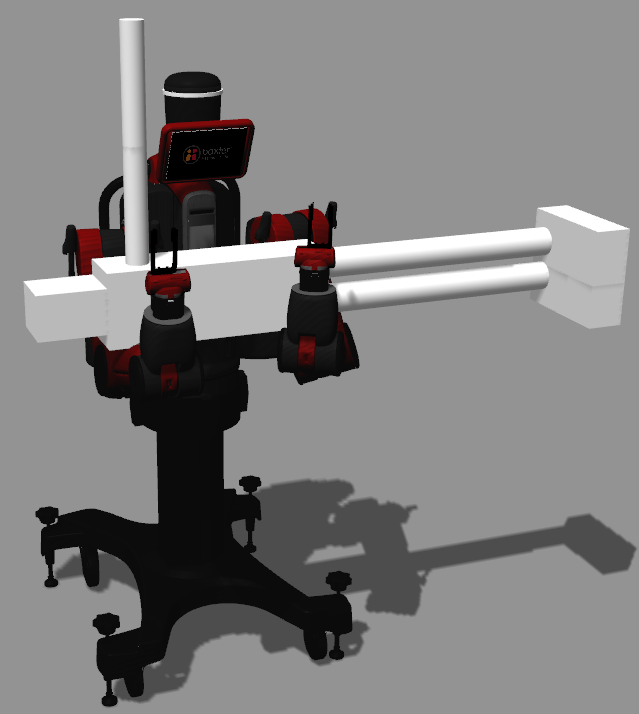}
    \caption{Horizontal slim humanoid}
    \label{fig:h_slim}
  \end{subfigure}
  \begin{subfigure}[b]{0.35\columnwidth}
    \includegraphics[height=3.5cm, trim={0cm 0cm 0cm 0cm}, clip]{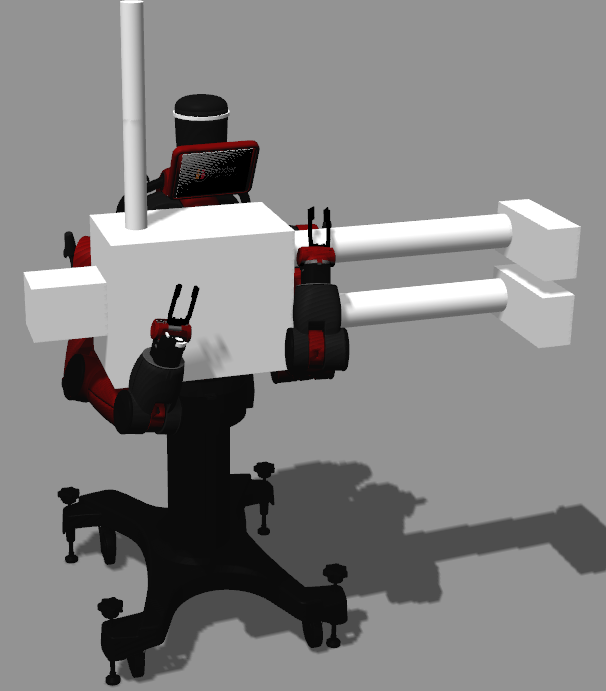}
    \caption{Horizontal stout humanoid}
    \label{fig:h_stout}
  \end{subfigure}

\caption{Holding examples for horizontal humanoid case. (a) is with the slim humanoid and (b) is with the stout humanoid.}
\label{fig:horizontal}
\end{figure}

In addition to holding the upright humanoid models, we applied the learned policy to a fixed floating humanoid as shown in Fig.~\ref{fig:holding}(f). Although the humanoid is spatially different from the upright model, our network was still able to tightly hold the humanoid by winding around the same links. Besides, using the linking $\Gamma_{\mathrm{H}}$ developed in Sec.~\ref{sec:statement}, we trained another policy and successfully applied it to hold horizontal humanoid models as demonstrated in Fig.~\ref{fig:robot}(b) and Fig.~\ref{fig:horizontal}. In addition to the robustness against differently shaped and scaled models, this implies that our formulation of the problem and the developed topological representation are flexible to the orientation of the humanoid model as well.

\begin{figure}[t]
\centering
\begin{subfigure}[b]{0.61\columnwidth}
  \includegraphics[height=3.5cm]{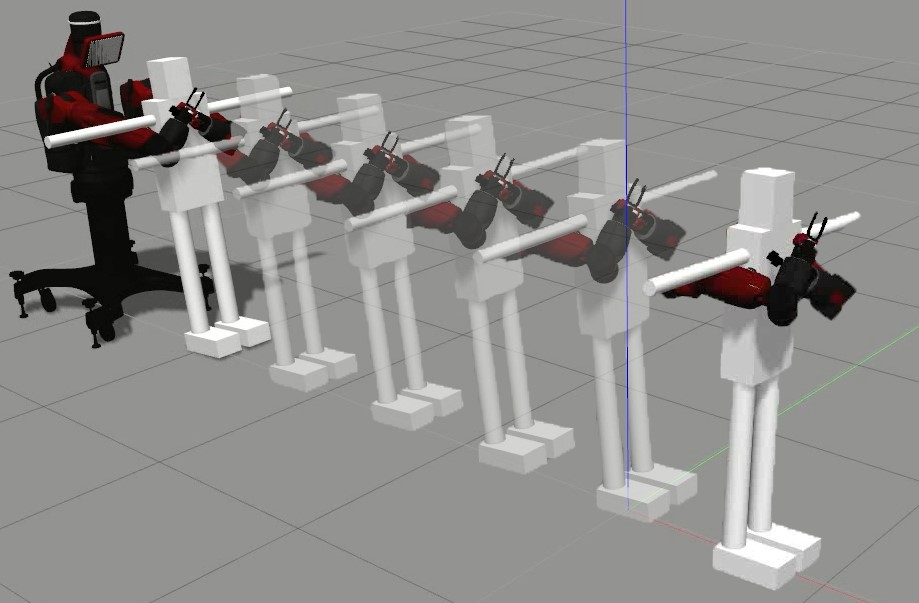}
  \caption{Dragging}
  \label{fig:drag}
\end{subfigure}
\begin{subfigure}[b]{0.33\columnwidth}
  \includegraphics[height=3.5cm]{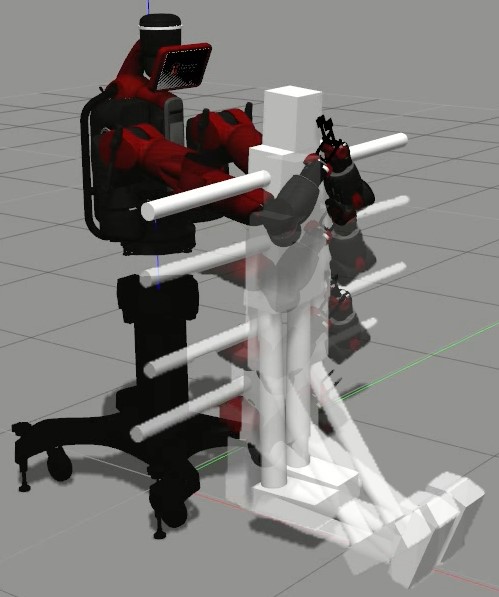}
  \caption{Lifting}
  \label{fig:lift}
\end{subfigure}
\caption{Dragging and lifting after holding is achieved.}
\label{fig:draglift}
\end{figure}

Moreover, once a holding is achieved, we tried to apply it to two different application cases based on the physical simulation in Gazebo, as shown in Fig.~\ref{fig:draglift}. For a standing humanoid, we moved the robot backwards to show that the achieved holding can stably pull the humanoid for transportation. When the humanoid is sitting on the floor, we show that the holding action can safely help it to stand up.

% \subsubsection{Dragging and Lifting}

% To show the stability of the subsequent manipulation after the holding, we let the robot move after it holds the humanoid to drag the humanoid away from the origin and to lift a fallen humanoid up. The dragging and lifting process are presented in Fig.~\ref{fig:draglift}, which is smooth and stable. This proves our holding is useful for manipulating and moving a humanoid or an object.

% \subsubsection{Executing in reality}

% To see if our policy can be applied to real world, we directly use the policy trained in simulator with a real robot. We read the arm link positions from the robot server and extract the humanoid skeleton positions using a Kinect depth sensor. Applied in reality without any fine-tuning, the holding policy still performs well, as is shown in Fig.~\ref{fig:real}. This proves that the state shift between the simulator and real world is trivial and the reality gap for our policy is small. It can be easily generalized to real robots.

\begin{figure}[t]
\centering
\includegraphics[width=0.5\columnwidth, trim={0cm 0cm 0cm 4cm}, clip]{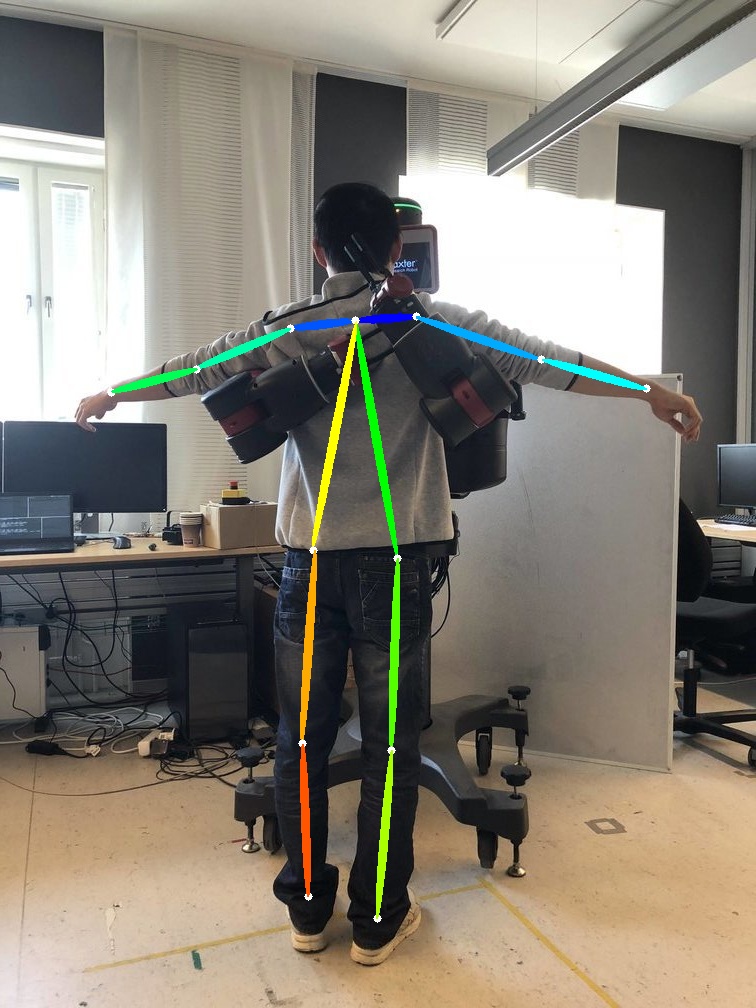}
\caption{Policy execution in reality: the human is perceived using a depth camera and the skeleton is extracted from the depth data.}
\label{fig:real}
% \vspace{-0.5cm}
\end{figure}

Lastly, we applied the policy trained in simulation directly to a real robot as in Fig.~\ref{fig:real}. The human was successfully held by the robot without requiring any extra tuning. This further shows one of the most important benefits of using topological representations that, since it is insensitive to geometries or perceptions, it can be easily transferred from simulation to reality.

%% file: images/evalu_writheinter.tex
\definecolor{mycolor1}{rgb}{0.00000,0.44700,0.74100}%
\begin{tikzpicture}

\begin{axis}[%
width=0.8\columnwidth,
height=0.3\columnwidth,
scale only axis,
xmode=log,
xmin=0,
xmax=2300,
xtick={10,100,500,2000},
xticklabels={{10},{100},{500},{2000}},
xminorticks=true,
xlabel={\#Episode},
title={(b) Evaluation},
ymin=0,
ymax=2.6,
ytick={  0, 0.5,   1, 1.5,   2, 2.5},
ylabel={$\Gamma_{\mathrm{U}}$},
axis x line*=bottom,
axis y line*=left,
legend style={at={(1,0.3)}, legend cell align=left, align=left, fill=none, draw=none}
]

\addplot[area legend, draw=none, fill=mycolor1, fill opacity=0.2, forget plot]
table[row sep=crcr] {%
x	y\\
10	0.264235674577377\\
20	0.397820161389662\\
30	0.875232024305404\\
40	0.281291201454123\\
50	-0.0883689369058236\\
60	0.850462232459974\\
70	0.642117705639585\\
80	1.03169731768133\\
90	0.635510550149585\\
100	1.31256575147059\\
110	1.2251078980211\\
120	0.902163592226034\\
130	1.40444697115441\\
140	1.215555305656\\
150	1.39636277992058\\
160	0.774485289752166\\
170	0.958676619025981\\
180	0.820150326328557\\
190	1.2321379249656\\
200	0.941578279164646\\
210	1.05366115953996\\
220	1.24546755271931\\
230	1.48808146729147\\
240	1.17209968315258\\
250	1.68324808292804\\
260	1.16439805546805\\
270	1.20885509531527\\
280	1.39812245725336\\
290	0.974603984398388\\
300	1.09804464242128\\
310	1.51278728783336\\
320	1.2719686439801\\
330	1.64584459952102\\
340	1.27301198122726\\
350	1.40344492395062\\
360	1.35071333312797\\
370	1.4931906724777\\
380	1.31024009618935\\
390	1.61908206679566\\
400	1.36246811596337\\
410	1.36800527790731\\
420	1.5851886365698\\
430	1.38896194832305\\
440	1.2709594105786\\
450	1.08062211902232\\
460	1.23724449082095\\
470	1.63715508158091\\
480	1.63853706536519\\
490	1.68686014246768\\
500	1.57113545876262\\
510	1.53616579595381\\
520	1.59566872347997\\
530	1.54935598263547\\
540	1.59398392074614\\
550	1.68339181689943\\
560	0.762524423374605\\
570	1.52282097582826\\
580	1.48838070693223\\
590	1.65426364104383\\
600	1.4895790726498\\
610	1.58861568291159\\
620	1.64495081380929\\
630	1.62614279082204\\
640	0.902705436127747\\
650	1.62674338412901\\
660	1.41176236651644\\
670	1.42268364740984\\
680	1.15996533657021\\
690	1.58543787562904\\
700	1.14713572279914\\
710	1.46430176293429\\
720	1.68573246718011\\
730	1.53503655881075\\
740	1.5609598607119\\
750	1.71460444867394\\
760	1.45451445771903\\
770	1.6993943294931\\
780	1.61556975901191\\
790	1.75868577949033\\
800	1.48839489774234\\
810	1.45142450434394\\
820	1.42734131342646\\
830	1.68017086934675\\
840	1.62797810247015\\
850	1.66585047311908\\
860	1.53867590620042\\
870	1.4643531750101\\
880	1.5180153878387\\
890	1.64047594527526\\
900	1.38517649108516\\
910	1.60863911254717\\
920	1.64520994670565\\
930	1.06811700422112\\
940	1.42800360844758\\
950	0.94555180486825\\
960	1.51834623461585\\
970	1.66063182697132\\
980	1.62671941160088\\
990	1.55250917862903\\
1000	1.66075629797345\\
1010	1.59322509134446\\
1020	1.52467132806687\\
1030	1.72076540509851\\
1040	1.6988882717114\\
1050	1.61916956614798\\
1060	1.3459553840113\\
1070	1.62362289139162\\
1080	1.4965705773089\\
1090	1.4634111571068\\
1100	1.73547311647565\\
1110	1.44518683496169\\
1120	1.36664259349568\\
1130	1.58936410394094\\
1140	1.4975192618505\\
1150	1.13981322680853\\
1160	1.51093820097396\\
1170	1.50753188446699\\
1180	1.51558809101453\\
1190	1.5626100973406\\
1200	1.57774150467305\\
1210	1.79532544867346\\
1220	1.48032983349517\\
1230	1.49127276970996\\
1240	1.65693633550854\\
1250	1.44154470732367\\
1260	1.55646408238164\\
1270	1.38717252606598\\
1280	1.67631049587853\\
1290	1.60278303378457\\
1300	1.55057010516465\\
1310	1.2902501448207\\
1320	1.58454345211148\\
1330	1.76528601384823\\
1340	1.3714228575136\\
1350	1.94596629024684\\
1360	1.00550966607868\\
1370	1.57525949470156\\
1380	1.48048716945017\\
1390	1.73376114668879\\
1400	1.5825526287754\\
1410	1.03324236139883\\
1420	1.49946330251521\\
1430	1.56476330037541\\
1440	1.71955204321739\\
1450	1.54029134618009\\
1460	1.69241251123208\\
1470	1.71919646744078\\
1480	1.67438213434348\\
1490	1.41873814478202\\
1500	1.38911292757609\\
1510	1.65776892203106\\
1520	0.92298088192378\\
1530	1.76050311847401\\
1540	1.54555611354901\\
1550	1.3330255109641\\
1560	1.30150526037397\\
1570	1.53774367986007\\
1580	1.62472727480439\\
1590	1.68573537994189\\
1600	1.50470413992415\\
1610	1.62022489495553\\
1620	1.32199761445214\\
1630	1.72798695262749\\
1640	1.52456654437186\\
1650	1.73370832411646\\
1660	1.67407423344794\\
1670	1.65661483185988\\
1680	0.825618687292635\\
1690	1.49049054324619\\
1700	1.03463057012629\\
1710	1.67943739349902\\
1720	1.68269234339054\\
1730	1.58641277590763\\
1740	1.50784014873242\\
1750	1.06800391399944\\
1760	1.40767199187111\\
1770	1.53398685851911\\
1780	1.51822792741215\\
1790	1.51406103381559\\
1800	1.53212681352344\\
1810	1.51340797612387\\
1820	1.7234699286615\\
1830	1.61708026216043\\
1840	1.46311260223205\\
1850	1.35605839641325\\
1860	1.54123375042772\\
1870	1.43391632867021\\
1880	1.34818768831277\\
1890	1.64899117926238\\
1900	1.64234629327711\\
1910	1.60058826344372\\
1920	1.53113632263849\\
1930	1.64336577053327\\
1940	1.02866730686095\\
1950	1.42408192788756\\
1960	1.65065745496576\\
1970	1.46904373130534\\
1980	1.65278232392869\\
1990	1.46272856943315\\
2000	1.40325932818174\\
2000	2.48330756237419\\
1990	2.39340401009435\\
1980	2.33609445922422\\
1970	2.38403608336814\\
1960	2.34976843286739\\
1950	2.64911191812711\\
1940	2.67682898403123\\
1930	2.25739295344295\\
1920	1.94393602125624\\
1910	2.22831605700941\\
1900	2.32339602278385\\
1890	2.18194428558202\\
1880	2.58209995843086\\
1870	2.4568588599203\\
1860	2.50267700431045\\
1850	2.3969755948084\\
1840	2.14103546995756\\
1830	2.28561739553939\\
1820	2.2287526869101\\
1810	2.37463349394599\\
1800	2.05007643412007\\
1790	2.3301814862461\\
1780	2.38761677846848\\
1770	2.32191359295732\\
1760	2.26734777270342\\
1750	2.5182753155833\\
1740	2.64110214309067\\
1730	2.25897072308276\\
1720	2.3050597346613\\
1710	2.42752883033759\\
1700	2.59844477698803\\
1690	2.35150373090102\\
1680	2.72479657712508\\
1670	2.33554440307322\\
1660	2.4016297312205\\
1650	2.33301814375339\\
1640	2.37741170981437\\
1630	2.18884967181144\\
1620	2.61566931756445\\
1610	1.99235547062265\\
1600	2.45796635013273\\
1590	2.27571653031367\\
1580	2.24827868491738\\
1570	2.60012147332869\\
1560	2.24013905264673\\
1550	2.40826379960194\\
1540	2.42234836773154\\
1530	2.15045966744214\\
1520	2.45357280792124\\
1510	2.35937895324852\\
1500	2.44762963821162\\
1490	2.55033571830691\\
1480	2.30802481932597\\
1470	2.21252922312737\\
1460	2.23082084604366\\
1450	2.14807091847915\\
1440	1.91832943827123\\
1430	2.3203528173803\\
1420	2.41231983779189\\
1410	2.5606929931485\\
1400	2.36372729117606\\
1390	2.02816298406013\\
1380	2.58315195589533\\
1370	2.09500789090222\\
1360	2.54070844890589\\
1350	2.16804770130536\\
1340	2.46976211318946\\
1330	2.19454103782867\\
1320	2.38393834427091\\
1310	2.46187760906152\\
1300	2.61397500029437\\
1290	2.39025758229603\\
1280	2.39584415271404\\
1270	2.30725706306797\\
1260	2.32692859422769\\
1250	2.5984987216542\\
1240	2.15373633014705\\
1230	2.11820214936462\\
1220	2.42560293582831\\
1210	2.25611583949804\\
1200	2.4741216077326\\
1190	2.30038497545986\\
1180	2.08361173100203\\
1170	2.22132193036724\\
1160	2.29885666219418\\
1150	2.53348355930255\\
1140	2.39420167275599\\
1130	2.61024132617932\\
1120	2.52722862226702\\
1110	2.42137631439914\\
1100	2.41672983429225\\
1090	2.44718755206766\\
1080	2.19972324304396\\
1070	2.42260570177046\\
1060	2.54010827659534\\
1050	2.35771459153595\\
1040	2.21776311489022\\
1030	2.37022358319445\\
1020	2.17361722790115\\
1010	2.38184876192263\\
1000	2.1091540239485\\
990	2.37967202566457\\
980	2.0747829307683\\
970	2.52955139213817\\
960	2.33539510653113\\
950	2.79126145267648\\
940	2.42233485484927\\
930	2.70870621741517\\
920	1.93050322742739\\
910	2.35987636997829\\
900	2.34168909193844\\
890	2.21888281472968\\
880	2.31679618442705\\
870	2.1704810232621\\
860	2.34848118072578\\
850	2.02595922244322\\
840	2.00617447682432\\
830	2.27856096408141\\
820	2.36670176456819\\
810	2.40103077334484\\
800	2.32867315542589\\
790	2.21840303527031\\
780	2.37577800753766\\
770	2.1021751240211\\
760	2.33773910852595\\
750	2.07768281340494\\
740	2.0916201752756\\
730	2.288706138627\\
720	2.14412270460602\\
710	2.14382474519092\\
700	2.48975342639173\\
690	2.14010713571836\\
680	2.61091155334766\\
670	2.443915475077\\
660	2.08164231609178\\
650	2.30706970729883\\
640	2.70846082611898\\
630	2.23808464001789\\
620	2.12068232814917\\
610	2.31527999230968\\
600	2.22907982497685\\
590	2.24202530998687\\
580	2.27300593032168\\
570	2.03964443339046\\
560	2.42644418487028\\
550	2.00670523832337\\
540	1.99185745702082\\
530	2.30555936875408\\
520	2.05891458925695\\
510	2.40890959410156\\
500	2.41252685453416\\
490	2.2229191025694\\
480	2.13637759641676\\
470	2.25729061134283\\
460	2.22696731401491\\
450	2.39786916615723\\
440	2.26400630871097\\
430	2.32761014621256\\
420	2.21716122348405\\
410	2.42227814348697\\
400	2.24230832336223\\
390	2.20120601860824\\
380	2.44299868812999\\
370	2.21998448389355\\
360	2.28562187493675\\
350	2.33616379082155\\
340	2.29621802981309\\
330	2.15425875415434\\
320	2.06079608476226\\
310	1.97842900703412\\
300	2.06830271699541\\
290	2.1623868028795\\
280	2.12815446617032\\
270	2.27078672814014\\
260	2.29005421546741\\
250	1.98327543823605\\
240	2.45421465013474\\
230	2.15396279930745\\
220	2.4485439363162\\
210	2.26519720581359\\
200	2.5578035721031\\
190	2.0527413679195\\
180	2.41167088661728\\
170	2.24843325070991\\
160	2.438568929523\\
150	1.93930992366098\\
140	2.28262533290581\\
130	2.40010317441557\\
120	2.12731825179823\\
110	2.18296296096918\\
100	2.10476794710848\\
90	2.37426634734205\\
80	2.22012742559063\\
70	2.34839298199116\\
60	1.86887084335472\\
50	1.94005130983468\\
40	1.79918299220099\\
30	1.08621921305656\\
20	1.00966493469237\\
10	0.809065224741907\\
}--cycle;
\addplot [color=mycolor1]
  table[row sep=crcr]{%
10	0.536650449659642\\
20	0.703742548041018\\
30	0.98072561868098\\
40	1.04023709682756\\
50	0.92584118646443\\
60	1.35966653790735\\
70	1.49525534381537\\
80	1.62591237163598\\
90	1.50488844874582\\
100	1.70866684928954\\
110	1.70403542949514\\
120	1.51474092201213\\
130	1.90227507278499\\
140	1.7490903192809\\
150	1.66783635179078\\
160	1.60652710963758\\
170	1.60355493486794\\
180	1.61591060647292\\
190	1.64243964644255\\
200	1.74969092563388\\
210	1.65942918267677\\
220	1.84700574451776\\
230	1.82102213329946\\
240	1.81315716664366\\
250	1.83326176058204\\
260	1.72722613546773\\
270	1.73982091172771\\
280	1.76313846171184\\
290	1.56849539363895\\
300	1.58317367970834\\
310	1.74560814743374\\
320	1.66638236437118\\
330	1.90005167683768\\
340	1.78461500552017\\
350	1.86980435738609\\
360	1.81816760403236\\
370	1.85658757818562\\
380	1.87661939215967\\
390	1.91014404270195\\
400	1.8023882196628\\
410	1.89514171069714\\
420	1.90117493002692\\
430	1.8582860472678\\
440	1.76748285964478\\
450	1.73924564258978\\
460	1.73210590241793\\
470	1.94722284646187\\
480	1.88745733089098\\
490	1.95488962251854\\
500	1.99183115664839\\
510	1.97253769502768\\
520	1.82729165636846\\
530	1.92745767569478\\
540	1.79292068888348\\
550	1.8450485276114\\
560	1.59448430412244\\
570	1.78123270460936\\
580	1.88069331862695\\
590	1.94814447551535\\
600	1.85932944881332\\
610	1.95194783761063\\
620	1.88281657097923\\
630	1.93211371541996\\
640	1.80558313112337\\
650	1.96690654571392\\
660	1.74670234130411\\
670	1.93329956124342\\
680	1.88543844495894\\
690	1.8627725056737\\
700	1.81844457459544\\
710	1.8040632540626\\
720	1.91492758589307\\
730	1.91187134871887\\
740	1.82629001799375\\
750	1.89614363103944\\
760	1.89612678312249\\
770	1.9007847267571\\
780	1.99567388327479\\
790	1.98854440738032\\
800	1.90853402658411\\
810	1.92622763884439\\
820	1.89702153899733\\
830	1.97936591671408\\
840	1.81707628964724\\
850	1.84590484778115\\
860	1.9435785434631\\
870	1.8174170991361\\
880	1.91740578613288\\
890	1.92967938000247\\
900	1.8634327915118\\
910	1.98425774126273\\
920	1.78785658706652\\
930	1.88841161081815\\
940	1.92516923164843\\
950	1.86840662877236\\
960	1.92687067057349\\
970	2.09509160955475\\
980	1.85075117118459\\
990	1.9660906021468\\
1000	1.88495516096098\\
1010	1.98753692663355\\
1020	1.84914427798401\\
1030	2.04549449414648\\
1040	1.95832569330081\\
1050	1.98844207884196\\
1060	1.94303183030332\\
1070	2.02311429658104\\
1080	1.84814691017643\\
1090	1.95529935458723\\
1100	2.07610147538395\\
1110	1.93328157468042\\
1120	1.94693560788135\\
1130	2.09980271506013\\
1140	1.94586046730324\\
1150	1.83664839305554\\
1160	1.90489743158407\\
1170	1.86442690741711\\
1180	1.79959991100828\\
1190	1.93149753640023\\
1200	2.02593155620282\\
1210	2.02572064408575\\
1220	1.95296638466174\\
1230	1.80473745953729\\
1240	1.9053363328278\\
1250	2.02002171448894\\
1260	1.94169633830466\\
1270	1.84721479456697\\
1280	2.03607732429628\\
1290	1.9965203080403\\
1300	2.08227255272951\\
1310	1.87606387694111\\
1320	1.98424089819119\\
1330	1.97991352583845\\
1340	1.92059248535153\\
1350	2.0570069957761\\
1360	1.77310905749229\\
1370	1.83513369280189\\
1380	2.03181956267275\\
1390	1.88096206537446\\
1400	1.97313995997573\\
1410	1.79696767727367\\
1420	1.95589157015355\\
1430	1.94255805887785\\
1440	1.81894074074431\\
1450	1.84418113232962\\
1460	1.96161667863787\\
1470	1.96586284528407\\
1480	1.99120347683473\\
1490	1.98453693154447\\
1500	1.91837128289386\\
1510	2.00857393763979\\
1520	1.68827684492251\\
1530	1.95548139295808\\
1540	1.98395224064028\\
1550	1.87064465528302\\
1560	1.77082215651035\\
1570	2.06893257659438\\
1580	1.93650297986088\\
1590	1.98072595512778\\
1600	1.98133524502844\\
1610	1.80629018278909\\
1620	1.9688334660083\\
1630	1.95841831221947\\
1640	1.95098912709312\\
1650	2.03336323393493\\
1660	2.03785198233422\\
1670	1.99607961746655\\
1680	1.77520763220886\\
1690	1.9209971370736\\
1700	1.81653767355716\\
1710	2.0534831119183\\
1720	1.99387603902592\\
1730	1.9226917494952\\
1740	2.07447114591154\\
1750	1.79313961479137\\
1760	1.83750988228726\\
1770	1.92795022573821\\
1780	1.95292235294032\\
1790	1.92212126003085\\
1800	1.79110162382176\\
1810	1.94402073503493\\
1820	1.9761113077858\\
1830	1.95134882884991\\
1840	1.80207403609481\\
1850	1.87651699561082\\
1860	2.02195537736908\\
1870	1.94538759429526\\
1880	1.96514382337182\\
1890	1.9154677324222\\
1900	1.98287115803048\\
1910	1.91445216022656\\
1920	1.73753617194736\\
1930	1.95037936198811\\
1940	1.85274814544609\\
1950	2.03659692300733\\
1960	2.00021294391658\\
1970	1.92653990733674\\
1980	1.99443839157646\\
1990	1.92806628976375\\
2000	1.94328344527797\\
};

\end{axis}
% \node[above,font=\scriptsize\bf] at (2.0, 3.0) {(b)Writhe number};

\end{tikzpicture}%

%% file: images/test_noisy.tex
\definecolor{mycolor1}{rgb}{0.63500,0.07800,0.18400}%
\definecolor{mycolor2}{rgb}{1.00000,0.65490,0.00000}%
\definecolor{mycolor3}{rgb}{0.46600,0.67400,0.18800}%
\definecolor{mycolor4}{rgb}{0.00000,0.44700,0.74100}%
\begin{tikzpicture}

\begin{axis}[%
width=0.5\columnwidth,
height=0.15\columnwidth,
at={(0\columnwidth,0\columnwidth)},
scale only axis,
xmin=0,
xmax=5,
xtick={0,1,2,3,4,5},
xlabel={\#Step},
ymin=0,
ymax=2.3,
ytick={  0, 0.5,   1, 1.5,   2},
ylabel={$\Gamma_{\mathrm{U}}$},
axis x line*=bottom,
axis y line*=left,
legend style={at={(1,0.65)}, legend columns = 2, legend cell align=left, align=left, fill=none, draw=none}
]

\addplot[area legend, draw=none, fill=mycolor1, fill opacity=0.1, forget plot]
table[row sep=crcr] {%
x	y\\
0	0.141200294167323\\
1	0.674730303776638\\
2	1.27935751403799\\
3	1.52632111063946\\
4	1.45356339641196\\
5	1.36081171634033\\
6	1.23433199729526\\
7	1.1904006066199\\
8	1.1556793076681\\
9	1.19099658581501\\
10	1.17520785645254\\
10	2.23408806022305\\
9	2.24990068183943\\
8	2.28085992564744\\
7	2.26855366475422\\
6	2.27085916765068\\
5	2.24689513663753\\
4	2.23086892813294\\
3	2.20562851165362\\
2	1.86840809664464\\
1	1.11202839538752\\
0	0.295259166739966\\
}--cycle;

\addplot[area legend, draw=none, fill=mycolor2, fill opacity=0.1, forget plot]
table[row sep=crcr] {%
x	y\\
0	0.128440680002893\\
1	0.759577881764135\\
2	1.26653312369122\\
3	1.51264403522993\\
4	1.47566846135293\\
5	1.40558634628491\\
6	1.36753863388162\\
7	1.33587392963292\\
8	1.27893508024273\\
9	1.20055288767817\\
10	1.15142961423444\\
10	2.34852670926816\\
9	2.3749720557056\\
8	2.38642300633144\\
7	2.42226892568113\\
6	2.47383097367282\\
5	2.46011326381935\\
4	2.42284448909453\\
3	2.30899222276783\\
2	1.90031126184242\\
1	1.12170112096062\\
0	0.313239038962154\\
}--cycle;

\addplot[area legend, draw=none, fill=mycolor3, fill opacity=0.1, forget plot]
table[row sep=crcr] {%
x	y\\
0	0.136822110113552\\
1	0.735012611052775\\
2	1.29749760027447\\
3	1.67621354690443\\
4	1.59866968146135\\
5	1.48350589487373\\
6	1.39493537681902\\
7	1.28405395722398\\
8	1.22011594206964\\
9	1.22511408466266\\
10	1.21602198013491\\
10	2.34808964811601\\
9	2.38209856240139\\
8	2.4383783338702\\
7	2.46410063769085\\
6	2.45347277442293\\
5	2.43546858774398\\
4	2.38181265416381\\
3	2.30497050655883\\
2	1.92011704390687\\
1	1.08815949495954\\
0	0.301507944538483\\
}--cycle;

\addplot[area legend, draw=none, fill=mycolor4, fill opacity=0.1, forget plot]
table[row sep=crcr] {%
x	y\\
0	0.135760948592158\\
1	0.738308205070525\\
2	1.32232535113641\\
3	1.68523144200498\\
4	1.65590214942406\\
5	1.51277304838426\\
6	1.41976550797395\\
7	1.33357741739335\\
8	1.26886988914225\\
9	1.25253971014361\\
10	1.20727715015588\\
10	2.34532189058324\\
9	2.35123286844569\\
8	2.37102524431903\\
7	2.36915595856799\\
6	2.38897557566597\\
5	2.39957135122212\\
4	2.28806322586685\\
3	2.26550167311309\\
2	1.90064952586799\\
1	1.09556172742313\\
0	0.314552291475373\\
}--cycle;
\addplot [color=mycolor4]
  table[row sep=crcr]{%
0	0.225156620033766\\
1	0.916934966246827\\
2	1.6114874385022\\
3	1.97536655755903\\
4	1.97198268764546\\
5	1.95617219980319\\
6	1.90437054181996\\
7	1.85136668798067\\
8	1.81994756673064\\
9	1.80188628929465\\
10	1.77629952036956\\
};
\addlegendentry{0}

\addplot [color=mycolor3]
  table[row sep=crcr]{%
0	0.219165027326017\\
1	0.911586053006159\\
2	1.60880732209067\\
3	1.99059202673163\\
4	1.99024116781258\\
5	1.95948724130886\\
6	1.92420407562098\\
7	1.87407729745741\\
8	1.82924713796992\\
9	1.80360632353203\\
10	1.78205581412546\\
};
\addlegendentry{0.1}

\addplot [color=mycolor2]
  table[row sep=crcr]{%
0	0.220839859482524\\
1	0.94063950136238\\
2	1.58342219276682\\
3	1.91081812899888\\
4	1.94925647522373\\
5	1.93284980505213\\
6	1.92068480377722\\
7	1.87907142765703\\
8	1.83267904328709\\
9	1.78776247169189\\
10	1.7499781617513\\
};
\addlegendentry{0.2}

\addplot [color=mycolor1]
  table[row sep=crcr]{%
0	0.218229730453645\\
1	0.893379349582081\\
2	1.57388280534131\\
3	1.86597481114654\\
4	1.84221616227245\\
5	1.80385342648893\\
6	1.75259558247297\\
7	1.72947713568706\\
8	1.71826961665777\\
9	1.72044863382722\\
10	1.70464795833779\\
};
\addlegendentry{0.3}

\end{axis}
\end{tikzpicture}%

%% file: includes/conclusion.tex
% !TEX root =  ../main.tex

\section{CONCLUSION}

In this work, we learned a motion policy that enabled WAM of a humanoid with close interaction between the humanoid's and the robot's bodies. We used a topology-based representation with \emph{Writhe matrix} and \emph{Laplacian coordinates} for reinforcement learning to achieve generalization and reactive behavior in dynamic scenarios.
%
%used the total linking value to indicate how well the holding is and formed the learning reward.
%
%We quantitatively compared the learning performance of using different representations and evaluated the learned policy by testing its success rate in the task execution.
Our results showed that this representation performed better than geometric state encoding in training and achieved a $99\%$ success rate in test. We also demonstrated the robustness and generalization of our policy by applying it in scenarios with unseen, different shape humanoids, floating humanoid, and with perception noise. In the qualitative evaluation, we showed that subsequent transporting was feasible by dragging the humanoid away or lifting it up. Further, we directly applied the policy learned in simulation on a real robot to verify that the policy can be easily transferred to reality.

In future work, we plan to assist the interaction with force sensors mounted on the robot's arms, in which case the robot would know about physical contacts with the holding targets and the policy would be able to learn a more comfortable way of holding the humanoid.

\section*{ACKNOWLEDGEMENT}
This work was supported by the HKUST SSTSP project RoMRO (FP802), HKUST IGN project IGN16EG09, HKUST PGS Fund of Office of Vice-President (Research \& Graduate Studies) and Knut and Alice Wallenberg Foundation \& Foundation for Strategic Research.